\documentclass{article} % For LaTeX2e

\RequirePackage[table]{xcolor}
\definecolor{B}{RGB}{200,230,250}
\definecolor{G}{RGB}{254,202,195}

\usepackage{iclr2025_conference,times}

\usepackage{amsmath,amsfonts,bm}

\def\eqref#1{equation~\ref{#1}}
\def\1{\bm{1}}

\DeclareMathAlphabet{\mathsfit}{\encodingdefault}{\sfdefault}{m}{sl}
\SetMathAlphabet{\mathsfit}{bold}{\encodingdefault}{\sfdefault}{bx}{n}

\usepackage{url}

\usepackage[utf8]{inputenc} % allow utf-8 input
\usepackage[T1]{fontenc}    % use 8-bit T1 fonts
\usepackage{hyperref}       % hyperlinks
\usepackage{url}            % simple URL typesetting
\usepackage{booktabs}       % professional-quality tables
\usepackage{amsfonts}       % blackboard math symbols
\usepackage{nicefrac}       % compact symbols for 1/2, etc.
\usepackage{microtype}      % microtypography
\usepackage{makecell}
\usepackage{graphicx}
\usepackage{wrapfig}
\usepackage{caption}
\usepackage{listings}

\addtocontents{toc}{\protect\setcounter{tocdepth}{-1}}

\title{Is Your Model Really A Good Math Reasoner? \\Evaluating Mathematical Reasoning with Checklist}

\author{%
Zihao Zhou$^{12}$\thanks{\begin{minipage}[t]{\textwidth}Equal contribution. Email: \texttt{zihao.zhou@liverpool.ac.uk; nlp2ct.shudong@gmail.com}\end{minipage}}\;\quad Shudong Liu$^{3*}$\quad Maizhen Ning$^{12}$\quad Wei Liu$^4$ \textbf{Jindong Wang}$^5$\quad \\
\textbf{Derek F. Wong}$^3$\quad \textbf{Xiaowei Huang}$^2$\quad \textbf{Qiufeng Wang}$^1$\quad \textbf{Kaizhu Huang}$^6$\\
    \textsuperscript{1}Xi’an Jiaotong-liverpool University  \quad \textsuperscript{2}University of Liverpool \quad \textsuperscript{3}University of Macau \\ \textsuperscript{4}HKUST \quad \textsuperscript{5}Microsoft Research Asia\quad \textsuperscript{6}Duke Kunshan University
}

\iclrfinalcopy % Uncomment for camera-ready version, but NOT for submission.
\begin{document}

\maketitle

\begin{abstract}
Exceptional mathematical reasoning ability is one of the key features that demonstrate the power of large language models (LLMs). How to comprehensively define and evaluate the mathematical abilities of LLMs, and even reflect the user experience in real-world scenarios, has emerged as a critical issue. Current benchmarks predominantly concentrate on problem-solving capabilities, presenting a substantial risk of model overfitting and fails to accurately measure the genuine mathematical reasoning abilities. In this paper, we argue that if a model really understands a problem, it should be robustly and readily applied across a diverse array of tasks.
To this end, we introduce \textbf{\textsc{MathCheck}}, a well-designed checklist for testing task generalization and reasoning robustness, as well as an automatic tool to generate checklists efficiently. \textsc{MathCheck} includes multiple mathematical reasoning tasks and robustness tests to facilitate a comprehensive evaluation of both mathematical reasoning ability and behavior testing. Utilizing \textsc{MathCheck}, we develop \textbf{\textsc{MathCheck-GSM}} and \textbf{\textsc{MathCheck-GEO}} to assess mathematical textual reasoning and multi-modal reasoning capabilities, respectively, serving as upgraded versions of benchmarks including GSM8k, GeoQA, UniGeo, and Geometry3K.
We adopt \textsc{MathCheck-GSM} and \textsc{MathCheck-GEO} to evaluate over 26 LLMs and 17 multi-modal LLMs, assessing their comprehensive mathematical reasoning abilities. 
  Our results demonstrate that while frontier LLMs like GPT-4o continue to excel in various abilities on the checklist, many other model families exhibit a significant decline.
  Further experiments indicate that, compared to traditional math benchmarks, \textsc{MathCheck} better reflects true mathematical abilities and represents mathematical intelligence more linearly, thereby supporting our design. Using \textsc{MathCheck}, we can also efficiently conduct informative behavior analysis to deeply investigate models. Finally, we show that our proposed checklist paradigm can easily extend to other reasoning tasks for their comprehensive evaluation.\footnote{Data and code can be found here: \url{https://mathcheck.github.io/}}

  % , promoting more comprehensive evaluation of reasoning abilit

  % Finally, we show that our proposed \textsc{MathCheck} paradigm can also easily be applied to other reasnoing tasks, constitute a significant stride towards a more comprehensive evaluation of reasoning ability

  % Finally, we show the potential of applying \textsc{MathCheck} paradigm to other reasoning tasks, constitute a significant stride towards a better reasoning benchmark paradigm.
  
  % We hope that our practice and observation can constitute a significant stride towards a more comprehensive evaluation of reasoning ability\footnote{Data and code can be found here: \url{https://anonymous.4open.science/r/MathCheck/}}.
\end{abstract}

\begin{figure*}[t]
    \centering
    \includegraphics[scale=0.44]{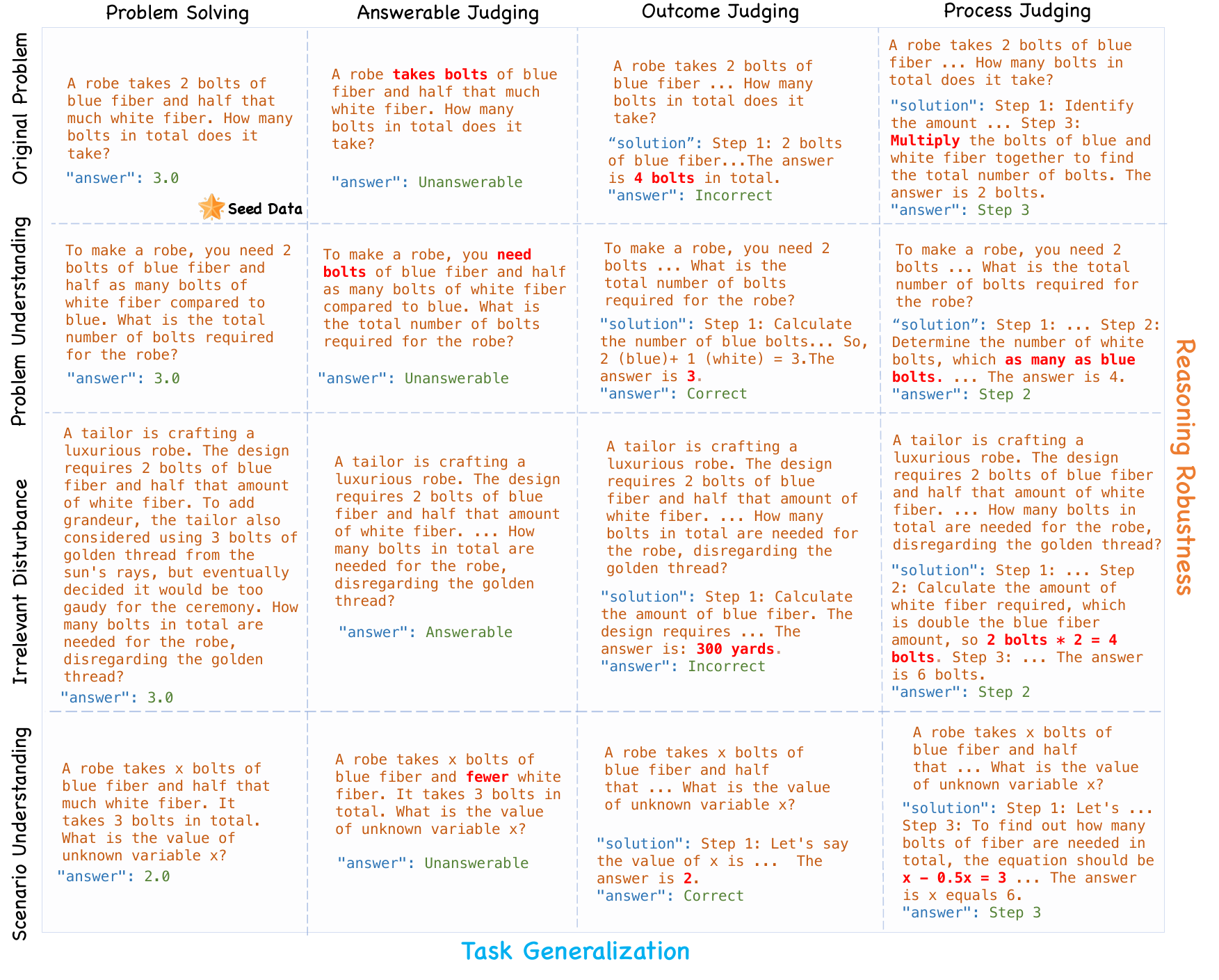}
    \caption{Overview of \textsc{MathCheck} design. The horizontal axis examines the task generalization of four math tasks while the vertical axis examines the reasoning robustness through four problem varieties. All data are generated from seed data, which is also from a mainstream benchmark dataset.}
    \label{fig:mathcheck}
\end{figure*}

% ----------------------------------------- Introduction -----------------------------------
\section{Introduction}
\label{introduction}
%Backgroud
    % Why study math ability
    % In the development of large language models (LLMs), mathematical reasoning ability stands as a crucial test-bed for LLMs' artificial general intelligence~\citep{}.
    The AI community has been placing significant emphasis on mathematical reasoning as a means to explore the upper limits of intelligence in large language models (LLMs)~\citep{achiam2023gpt,team2023gemini,Meta-llama,jiang2024mixtral,wei2022chain,trinh2024solving,romera2024mathematical} and multi-modal large language models (MLLMs)~\citep{GPT4V,lu2023mathvista}. 
    %what the reseacher work on
    A large number of efforts have been made on how to enhance (M)LLMs' mathematical reasoning abilities. In pre-training, \citet{wang2023generative,shao2024deepseekmath,lin2024rho,zhang2024autonomous} studied the impact of the quality of mathematical corpus; in post-training, \citet{yue2023mammoth,yu2023metamath,li2024common} augmented a huge number of synthetic data, and then developed supervised fine-tuning (SFT) for math problem-solving. Recently, \cite{luong2024reft} and \cite{sun2024easy} explored variants of reinforcement learning (RL) for further improvements.

%Motivation
    %how to evaluate
    To guarantee the high mathematical reasoning ability has been reached, it is crucial to fairly evaluate models' performance.
    Current mainstream methods rely on the performance across math problem-solving tasks of varying difficulty levels, such as GSM8k~\citep{cobbe2021training} of elementary level, MATH~\citep{hendrycks2021measuring} of high school level, and TheromQA~\citep{chen2023theoremqa} of university level. 
    % Recently, some harder or more diverse math datasets be proposed to enrich the evaluation~\cite{oly,..,liu2024mathbench,lu2023mathvista}. 
    Recently, some mathematical datasets that are more challenging, diverse, and multi-modal have been proposed to enhance the mathematical evaluation~\citep{he2024olympiadbench,liu2024mathbench,lu2023mathvista,zhang2024mathverse}.
    %issue
    However, these current evaluation methods focus on \emph{individual} tasks (most of which are problem-solving) and robustness tests for each problem. In other words, they do not provide comprehensive guidance on whether LLMs really achieve mathematical reasoning ability. 
    %our insight
    In this paper, we argue that: \textit{if a model really understands a problem, it should work robustly across various tasks about this problem. } Therefore, it is necessary to evaluate models by multi-tasks with diverse robustness test. Through such investigation, the real reasoning ability of a model can be comprehensively evaluated. As a result, we can also perform detailed behavior tests on models~\citep{ribeiro2020beyond}.

%Method
    %overview
    Drawing motivations from this insight, we introduce \textbf{\textsc{MathCheck}}, a well-designed checklist for testing task generalization and reasoning robustness.
    \textsc{MathCheck} includes general mathematical reasoning tasks and diverse robustness testing types to facilitate a comprehensive evaluation of mathematical reasoning ability and reasoning behavior testing.
    %MathCheck design
    As shown in Figure~\ref{fig:mathcheck}, horizontally, we examine the task generalization including problem solving, answerable judging, outcome judging, and process judging. Vertically, we test the reasoning robustness through the original problem and its three robustness variants consisting of problem understanding, irrelevant disturbance, and scenario understanding. The data of each cell in the checklist corresponds to a specific type of robustness test and task form. 
    %how to generate
    To facilitate the construction of checklist, we propose an (M)LLMs-driven generation framework to automatically generate this data. 
    Figure~\ref{fig:generation} illustrates the \textsc{MathCheck} data collection process, where the seed solving problem is firstly rewritten to its robustness problems, next all generated solving data are utilized to construct other task forms. %Specifically, we apply LLMs to generate new data because of their powerful rewritten capability and generalization~\citep{li2024gsm}.

%Mathcheck-Gsm and Geo
    %Gsm
    Utilizing \textsc{MathCheck}, we propose \textbf{\textsc{MathCheck-GSM}}, a \textsc{MathCheck} dataset generated from GSM8k~\citep{cobbe2021training}. It contains a total of 3,096 high-quality samples consisting of 129 groups checklist matrix, which can be used to evaluate mathematical textual reasoning ability comprehensively. 
    %Geo
    Besides, acknowledging the community's focus on multi-modal reasoning capabilities, we further propose \textbf{\textsc{MathCheck-GEO}} to evaluate the multi-modal geometry reasoning ability. Generated from GeoQA~\citep{chen2021geoqa}, UniGeo~\citep{chen2022unigeo}, and Geometry3K~\citep{lu2021inter}, it contains a total of 1,440 samples with a checklist matrix of 60 groups. 
    It is noteworthy that the construction pipeline of \textsc{MathCheck} can be applied to most mathematical datasets to dynamically establish a comprehensive and flexible evaluation benchmark, thereby mitigating data contamination~\citep{zhou2023don,zhu2024dyval,zhu2024dyval2}.
    
%Findings
    We conduct extensive experiments on 26 LLMs and 17 MLLMs including different scales, API-base and open source, generalist and mathematical models.
    %Main result
    We find that frontier LLMs like GPT-4o continue to achieve superior performance in our \textsc{MathCheck}, but many other model families exhibit a significant decline.
    % Comparing the excellent performance on original problem solving, Mathematical/reasoning specific models significantly become worse in other tasks or robustness variants. 
    Further experiments indicate that compared to solving original problems which is the paradigm of mainstream benchmark, our \textsc{MathCheck} evaluation aligns more accurately with the genuine mathematical reasoning ability of the model. Utilizing \textsc{MathCheck}, we extensively analyze the models' behaviors including training on massive solving data, reasoning consistency, performance on different complexity problems and applying different prompting technologies. Finally, we show the potential of applying \textsc{MathCheck} paradigm to other reasoning tasks such as commonsense reasoning and code generation, promoting more comprehensive evaluation of reasoning ability.
\section{\textsc{MathCheck}}
\textsc{MathCheck} is a well-designed checklist that includes general mathematical reasoning tasks and diverse robustness testing types for comprehensive evaluation, as well as a tool to automatically generate a large number of test cases in the manner of checklist. 
In our checklist, various mathematical tasks are arranged in rows to assess task generalization, whereas diverse variants of mathematical problems are placed in columns to evaluate reasoning robustness.
% In our checklist, various mathematical tasks on the rows to test task generalization and diverse variants of mathematical problems on the columns to test reasoning robustness respectively.
We will elaborate on the task types in Section~\ref{secTG}, problem variants in Section~\ref{secRR}, and how we construct checklist data in Section~\ref{sectool}.

\subsection{Task Generalization}\label{secTG}
Testing models across different tasks on the same domain not only offers a comprehensive and profound evaluation of their capabilities~\citep{frank2023baby} but also caters to the practical demands and complexities of real-world applications~\citep{ji2023systematic}. In \textsc{MathCheck}, we incorporate four math tasks including Problem Solving, Answerable Judging, Outcome Judging, and Process Judging.

% \paragraph{Solving}
\textbf{Problem Solving.}
In this task, we ask the model to solve a given math problem. As the most widely used method to test mathematical reasoning ability in contemporary research~\citep{cobbe2021training,hendrycks2021measuring}, it necessitates the model to analyze the problem, recall and apply appropriate math knowledge, and finally conclude reasoning results.

\textbf{Answerable Judging.}
Given a math problem, models need to determine whether the problem provides sufficient information to answer the question. This task requires the model to analyze the question, then identify the essential conditions required for solving this question, subsequently verify whether these conditions are provided within the problem statement. Previous works utilized it to examine whether the model is a reasoner with critical thinking instead of a random parrot~\citep{li2024gsm,sun2024benchmarking,ma2024large}.

\textbf{Outcome Judging.}
Given a math problem and one of its solutions, let the model determine whether the final answer of the given solution is correct. Outcome-Judging is a coarse-grained judgment of solutions since the model only focuses on the correctness of the final answer. Researchers often apply the outcome-judging ability of models to verify the correctness of augmented data~\citep{tang2024mathscale} and provide outcome rewards in reinforcement learning~\citep{luong2024reft}.

\textbf{Process Judging.}
Given a math problem along with its wrong solution,  the model is required to identify the step where the errors begin. Compared with the outcome-judging, the process-judging task is a more fine-grained judgment on the solution, which demands the model to judge step by step until the wrong step is located. It can help to debug the given wrong solution.

\subsection{Reasoning Robustness}\label{secRR}
A model that truly understands the inherent mathematical logic of a problem will exhibit reasoning robustness to diverse variations of this problem~\citep{stolfo2023causal}. Motivated by this, we utilize four problem forms including the original problem and its three rewritten variants to examine the reasoning robustness of models.

\textbf{Original Problem.}
It is the seed problem of other reasoning robustness variants. %adversarial problems. 
At a minimum functionality test, it can check whether the model has the basic mathematical capabilities when no modifications have been made.

\textbf{Problem Understanding.}
It refers to transforming the original problem into a new one that uses different wording or different sentence structures but does not change the mathematical logic of its original version~\citep{patel2021nlp,zhou2024mathattack,li2024gsm}. It pays more attention to semantic robustness, and aims to examine whether models can correctly reason when dealing with different descriptions of the same mathematical logic.

\textbf{Irrelevant Disturbance.}
It refers to inserting irrelevant conditions that are related to the topic of the original question, but have no impact on the final answer. Previous studies have disclosed that large language models are easily distracted by such perturbations~\citep{shi2023large}. It needs the model to distinguish which conditions are necessary and which are irrelevant to the problem.

\textbf{Scenario Understanding.}
When models comprehend the scenario of a math problem and its underlying logic, they should be able to solve other questions within that scenario~\citep{liu2021roda, yu2023metamath, zhou-etal-2023-learning-analogy}. Therefore, we alter the original question to evaluate whether a model has a comprehensive understanding of the scenario. For example, as shown in Figure~\ref{fig:mathcheck}, we ask the question ``the number of blue bolts" instead of ``the number of total bolts".

\begin{figure*}[t!]
    \centering
    \includegraphics[scale=0.52]{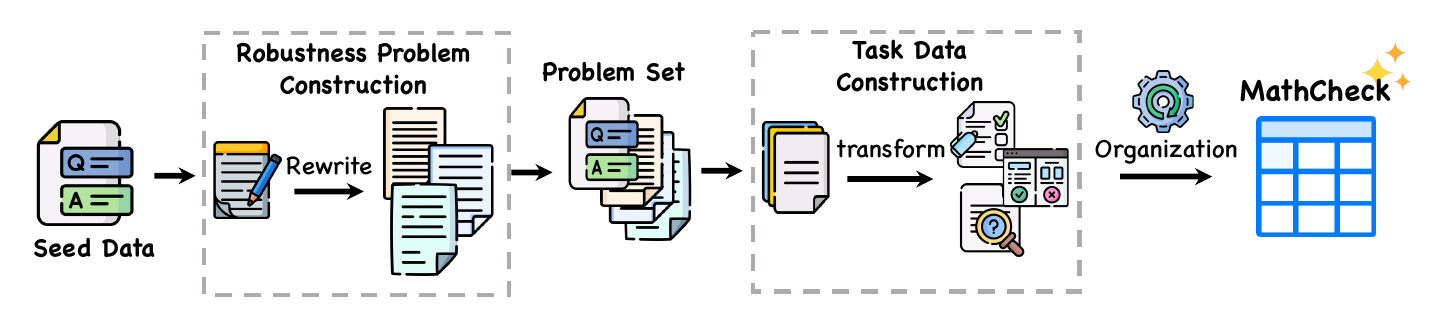}
    \caption{\textsc{MathCheck} generation pipeline. %First, we rewrite seed data into their robustness varieties, getting robustness problems set. Subsequently, we will transform each problem into multiple mathematical tasks about this problem. Finally,  we check all data manually, forming \textsc{MathCheck}.
    }
    \label{fig:generation}
\end{figure*}

% ----------------------------- Checklist Construction -----------------------------
\subsection{Checklist Construction}\label{sectool}
Creating \textsc{MathCheck} data is a labor-intensive and time-consuming process. 
% At the same time, we want everyone can build their own checklist based on their own data.
The advent of LLMs has introduced a new level of flexibility and quality to generate mathematical content~\citep{norberg2023rewriting,li2024gsm}. Therefore, we employ (M)LLMs (e.g., GPT-4-Turbo in our experiments) as engines to automatically generate our \textsc{MathCheck} data. 
%Therefore we propose an (M)LLMs-driven generation framework to automatically generate this data. 
The data construction pipeline is shown in Figure~\ref{fig:generation}. Users first assemble a collection of math problems with labels as seed data. Second, (M)LLMs initially rewrite these problems into their robustness varieties to make up the robustness problem set. Third, each problem in this set will be extended to construct multiple mathematical tasks about this problem. Finally, all data are manually checked to form \textsc{MathCheck} dataset correctly. 

%\subsubsection{Generation Process}
%Most prior methods rely on heuristic rules to rewrite mathematical problems~\citep{norberg2023rewriting,kumar2022practice}, but they are limited by poor adaptability and lack of diversity. 
% The emergence of LLMs makes mathematical generation flexible and high-quality~\citep{norberg2023rewriting,li2024gsm}, we apply (M)LLMs~\footnote{We use GPT-4-Turbo as our generation model.} as engines to generate our data.
%The advent of LLMs has introduced a new level of flexibility and quality to generate mathematical content~\citep{norberg2023rewriting,li2024gsm}. Therefore, we employ (M)LLMs (e.g., GPT-4-Turbo in our experiments) as engines to generate our \textsc{MathCheck} data.  

Based on the seed data, we automatically generate another three robustness problems as shown in the first column of Figure~\ref{fig:mathcheck}. 
\textit{Problem Understanding} and \textit{Irrelevant Disturbance} are the tasks of rewriting problems without altering the final answer. Hence, we prompt the model to rewrite our math problems while maintaining the original answer. For \textit{Scenario Understanding}, we first extract a variable from the problem as a new answer, then prompt the model to change the question based on the extracted variable. Once we obtain the four robustness reasoning problems of the solving task, we rewrite them respectively to construct multiple tasks, including \textit{Answerable Judging}, \textit{Outcome Judging} and \textit{Process Judging} as shown in the corresponding row of Figure~\ref{fig:mathcheck}. For the \textit{Answerable Judging} task, we prompt the model to eliminate a condition from the original problem which is crucial for solving it to obtain an unanswerable problem. For \textit{Outcome Judging} task, we ask the model to solve the problem and acquire candidate solutions, then these solutions are labeled (Correct or Incorrect) according to the final answer.  For \textit{Process Judging} task, we apply the solution rewritten ability of (M)LLMs to construct process-judging data. Specifically, given a problem along with its correct solution, we prompt the model to make mistakes from the given steps and results in a wrong answer. In such a way, we can get a wrong solution while its mistake steps remain simultaneously. 
All of our prompts are listed in Appendix~\ref{secGenPrompt}.

% -------------------------------------- Experiments ---------------------------------------

\section{Experiments}
\subsection{Datasets}
\label{Datasets}
% We use \textsc{MathCheck} to comprehensively measure the mathematical textual reasoning capability as well as the mathematical multi-modal reasoning capability. Therefore, we introduce \textsc{MathCheck-GSM} and \textsc{MathCheck-GEO}.

We use \textsc{MathCheck} to comprehensively measure the mathematical reasoning ability across textual and multi-modal settings. Consequently, two benchmarks \textsc{MathCheck-GSM} and \textsc{MathCheck-GEO} are introduced.

\textbf{\textsc{MathCheck-GSM}} is a \textsc{MathCheck} dataset generated from GSM8k~\citep{cobbe2021training}. We choose GSM8k as the seed benchmark since 
(1) it is most widely used for evaluating mathematical textual reasoning capability. 
(2) we aim to determine whether advanced models are genuinely capable of reasoning at the grade school level.
We first collect a test-mini set of GSM8k, which includes 129 problems sampled evenly according to the difficulty\footnote{We define the difficulty according to the number of reasoning steps of its answers (2 steps to 8 steps)}. Subsequently, we generate 129 \textsc{MathCheck} style groups, totaling 3,096 high-quality samples by \textsc{MathCheck}. It can be used to evaluate the real mathematical reasoning ability of LLMs on GSM8k-level problems. A group of \textsc{MathCheck-GSM} case problems are listed in Appendix~\ref{case_gsm}.

\textbf{\textsc{MathCheck-GEO}} is a dataset for geometry problems, which is the representative task for evaluating multi-modal reasoning capability. First, we collect seed geometry problems from GeoQA~\citep{chen2021geoqa}, UniGeo~\citep{chen2022unigeo}, and Geometry3K~\citep{lu2021inter}, containing 60 problems in both English and Chinese. Subsequently, we generate 60 \textsc{MathCheck} style groups, totaling 1,440 high-quality samples. Notably, this is the first geometry problem dataset involving answerable, outcome, and process judgment tasks. 
\textsc{MathCheck-GEO} gives research community a harder and multi-modal \textsc{MathCheck} style dataset, as well as showing the extensibility of \textsc{MathCheck}.
A group of \textsc{MathCheck-GEO} case problems are shown in Appendix~\ref{case_geo}.

All datasets are checked with meticulous manual validation to ensure high quality and reliability. To this end, we recruited three graduate students who underwent training tailored to the requirements of our research. This rigorous verification process not only enhances the quality of our data but also reinforces the validity of our findings. Finally, our automatic data generation pipeline can achieve an average pass rate of 84.61\% (Appendix~\ref{secPR}). The detailed data statistics and quality discussion of our checklist are reported in Appendix~\ref{datasta}.

% ----------------- Table: Model Performance on Tasks - GSM -----------------

\begin{table*}[t!]
  \centering
    \caption{Model performance on \textsc{MathCheck-GSM}.
     \textbf{PS}: \textbf{P}roblem \textbf{S}olving, 
      \textbf{AJ}: \textbf{A}nswerable \textbf{J}udging,  
      \textbf{OJ}: \textbf{O}utcome \textbf{J}udging, 
      \textbf{PJ}: \textbf{P}rocess \textbf{J}udging,
      \textbf{OP}: \textbf{O}rigianl \textbf{P}roblem, 
      \textbf{PU}: \textbf{P}roblem \textbf{U}nderstanding,
      \textbf{ID}: \textbf{I}rrelevant \textbf{D}isturbance, 
      \textbf{SU}: \textbf{S}cenario \textbf{U}nderstanding. Each score is the average score of related units. For example, 'All' means all units, 'PS'  includes solving units on four problem types, 'OP' includes original problems on four tasks units.
      }
  \scalebox{0.92}{
  \begin{tabular}{l|c|cccc|cccc}
    \toprule
    Models                      & All               & PS     & AJ    & OJ  & PJ        & OP   & PU   & ID    & SU   \\
    \midrule    % Close Model
    \rowcolor[gray]{0.88}
    \multicolumn{10}{c}{\textit{Generalist Models}} \\ 
    \midrule
    O1-preview              & \cellcolor{B}93.2             & 91.3   & 94.0  &\cellcolor{B}93.2 & \cellcolor{B}94.1      & \cellcolor{B}95.6 & 93.4 & 90.5  & \cellcolor{B}93.1  \\
    O1-mini              & 92.7             & 93.6   & 95.0  & 88.9 & 93.6      & 95.5 & \cellcolor{B}94.2 & 91.0  & 90.5  \\ 
    GPT-4o                &92.0 & \cellcolor{B}95.0   & 95.0  & 90.1 & 87.8      & 94.6 & 91.6 & \cellcolor{B}92.0  & 89.6  \\
    GPT-4o-mini              & 87.2             & 90.1   & 89.6  & 88.6 & 80.4      & 88.9 & 89.4 & 85.6  & 85.1  \\ 
    GPT-4-Turbo-20240409            & 90.9             & 93.8   & \cellcolor{B}95.9  & 87.8  & 86.0     & 93.8 & 90.4 & 90.8  & 88.6  \\
    GPT-3.5-Turbo               & 61.4             & 73.5   & 64.3  & 48.3 & 59.5      & 65.4 & 64.6 & 60.1  & 55.4  \\    
    Gemini-1.5-Pro              & 86.3             & 88.6   & 89.5  & 87.6 & 75.0      & 88.0 & 90.2 & 85.0  & 82.0  \\
    Claude-3.5-sonnet-20240620    & 90.2             & 94.8   & 95.3  & 90.9 & 79.9      & 92.5 & 92.1 & 89.9  & 86.3  \\
    Claude-3-opus-20240229      & 83.5             & 81.6   & 92.0  & 78.7 & 81.8      & 86.3 & 85.6 & 81.9  & 80.3  \\
    Claude-3-sonnet-20240229    & 75.0             & 77.9   & 88.9  & 65.1 & 68.0      & 76.5 & 77.8 & 73.7  & 71.9  \\
    Claude-3-haiku-20240229     & 57.5             & 79.7   & 49.9  & 44.3 & 56.0      & 61.9 & 62.4 & 55.9  & 49.6  \\
    \midrule    % Open Model
    Llama-3.1-70B-Instruct        &\cellcolor{B}90.5 &\cellcolor{B}95.2   &\cellcolor{B}95.3 &\cellcolor{B}89.4
    & \cellcolor{B}82.2 &\cellcolor{B}93.3 &\cellcolor{B}91.2 &\cellcolor{B}89.8  &\cellcolor{B}87.7  \\
    Llama-3-70B-Instruct        
    & 84.7 & 90.1  & 87.5  & 84.6 &76.7      & 87.7 & 86.7 & 84.7  & 79.9  \\
    DeepSeek V2          & 82.2             & 86.8   & 82.6  & 82.5 & 76.9   & 85.1 & 84.4 & 83.5  & 75.9  \\
    Mixtral 8 x 7B-Instruct     & 59.9             & 56.0   & 58.1  & 63.9 & 61.6      & 62.8 & 61.5 & 58.8  & 56.4  \\
    Mixtral 8 x 7B-Base         & 44.7             & 40.9   & 50.8  & 51.8 & 35.3      & 50.6 & 47.8 & 41.2  & 39.1  \\
    Qwen1.5-72B-Chat            & 50.6             & 71.1   & 64.2  & 31.9 & 35.1      & 57.0 & 51.1 & 43.6  & 50.6  \\
    \midrule
    Phi-3-Medium-4K-Instruct&\cellcolor{B}72.0&\cellcolor{B}89.7   &\cellcolor{B}70.8  &63.2 &\cellcolor{B}64.1      &\cellcolor{B}77.6 & \cellcolor{B}78.7 &\cellcolor{B}71.1  & 60.4  \\
    Phi-3-Mini-4K-Instruct  & 64.1              & 71.3   & 64.5  & 62.9 & 57.6      & 68.5 & 66.6 & 61.2  & 60.0 \\
    Llama-3.1-8B-Instruct          
    & 71.0 & 76.9 & 65.8  &\cellcolor{B}77.2 & 64.0 & 74.6 & 73.6 & 66.0  & \cellcolor{B} 69.6  \\
    Llama-3-8B-Instruct          & 64.2             & 68.6   & 61.4  & 64.9 & 61.8      & 67.8 & 68.8 & 62.9  & 57.1  \\
    ChatGLM3-6B                 & 36.5             & 32.6   & 41.7  & 50.1 & 21.7      & 39.7 & 35.9 & 31.3  & 39.1  \\
    \midrule
    \rowcolor[gray]{0.88}
    \multicolumn{10}{c}{\textit{Mathematical Models}} \\
    \midrule    % Math Model
    DeepSeek-Math-7B-RL         &\cellcolor{B}50.7  &\cellcolor{B}79.5   & 50.0  &\cellcolor{B}45.1 &28.1     &\cellcolor{B}53.3 & 51.2 & \cellcolor{B}47.5  &\cellcolor{B}50.6  \\
    DeepSeek-Math-7B-Instruct   & 50.2              & 70.0   & \cellcolor{B}64.8  & 40.4 & 25.8      & 51.6 &\cellcolor{B}54.4 & 45.8  & 49.2  \\
    DeepSeek-Math-7B-Base       & 44.0              & 49.8   & 51.5  & 44.0 & 30.8      & 49.0 & 46.0 & 37.0  & 44.1  \\
    MetaMath-LLama2-70B         & 45.7              & 70.0   & 35.7  & 45.3 &\cellcolor{B}31.6      & 49.9 & 51.5 & 43.4 & 37.8  \\
    \bottomrule
  \end{tabular}}
\label{table_1}
\end{table*}

% ----------------- Table: Model Performance on Tasks - Geometry -----------------

\begin{table*}[t!]
  \centering
      \caption{Model performance on \textsc{MathCheck-GEO}.}
  \scalebox{0.92}{
  \begin{tabular}{l|c|cccc|cccc}
    \toprule
    % \multicolumn{2}{c}{Part}                   \\
    Models                      & All              & PS     & AJ    & OJ    & PJ       & OP   & PU   & ID    & SU   \\
    % \multicolumn{10}{c}{General Models} \\
    \midrule    % Close Model
    \rowcolor[gray]{0.88}
    \multicolumn{10}{c}{\textit{Generalist Models}} \\ 
    \midrule
    GPT-4o                      &\cellcolor{G}65.3 &\cellcolor{G}57.5   &\cellcolor{G}75.5  &\cellcolor{G}69.5 & 58.8 &\cellcolor{G}65.2 &\cellcolor{G}67.0 &\cellcolor{G}64.3  &\cellcolor{G}64.8  \\
    GPT-4o-mini         & 59.0      & 50.8    & 69.8  & 61.4 & 53.8      & 61.9 & 62.0 & 54.1  & 57.8  \\    
    GPT-4-Turbo-20240409            & 61.7             & 51.3   & 72.3  & 64.0 & 59.2      & 63.2 & 62.9 & 61.7  & 58.9  \\
    GPT-4-Vision-Preview        & 60.0             & 46.7   & 71.1  & 63.6 & 58.8      & 59.3 & 62.8 & 57.8  & 60.2  \\
    Gemini-1.5-Pro              & 58.7             & 47.5   & 67.4  & 55.0 &\cellcolor{G}64.6      & 62.3 & 58.6 & 57.1  & 56.9  \\
    Gemini-1.5-Flash            & 56.8             & 45.0   & 75.1  & 50.6 & 56.7      & 56.8 & 59.7 & 53.8  & 57.1  \\
    Claude-3.5-sonnet-20240620              & 58.7             & 54.2    & 71.0  & 53.0 & 56.7      & 59.9 & 63.8 & 54.3  & 56.8  \\
    Claude-3-opus-20240229      & 47.2             & 34.2   & 60.6  & 46.7 & 47.5      & 47.2 & 49.1 & 42.4  & 50.2  \\
    Claude-3-sonnet-20240229    & 49.9             & 35.8   & 59.0  & 51.6 & 52.9      & 51.2 & 53.0 & 44.7  & 50.4  \\
    Claude-3-haiku-20240307     & 36.7             & 27.9   & 41.3  & 41.7 & 35.8      & 39.2 & 38.8 & 33.3  & 35.4 \\

    \midrule    % Open Model
    QWen2-VL-72B-Instruct                & \cellcolor{G}61.4             & \cellcolor{G}60.0    & 53.1  & \cellcolor{G}61.3 & \cellcolor{G}71.3      & \cellcolor{G}69.0 & \cellcolor{G}62.4 & \cellcolor{G}58.0  & \cellcolor{G}56.4  \\
    QWen2-VL-7B-Instruct                 & 42.1             & 35.8    & 49.4  & 46.4 & 36.7      & 40.9 & 45.6 & 41.7  & 40.0  \\
    InternVL-1.5-Chat           &37.6 &22.1   & \cellcolor{G}54.9  & 46.8 &26.7    &42.9 &34.8 &37.3  &35.5  \\
    MiniCPM-Llama3-V-2.5	&37.3	&37.5	&38.1	&45.0	&28.8	&37.4	&45.0	&35.2	&31.6 \\
    LLaVA-1.6-Mistral-7B-Instruct&31.8&10.0	&38.8	&51.2	&27.1	&33.8	&35.5	&28.4	&29.2 \\
    Phi-3-Vision-128k-Instruct  & 29.6             & 12.9   & 35.0  &48.6 & 22.9  &32.6 &31.8 &28.2  & 26.0  \\
    CogVLM2-Llama3-Chat-19B     & 24.6             & 7.9    & 26.4  & 46.3 & 17.9      & 27.2 & 28.0 & 22.4  & 20.9  \\
    \bottomrule
  \end{tabular}}
  \label{table_2}
\end{table*}

% ----------------------- Experimental Setup ------------------------

\subsection{Experimental Setup}
\label{Experimental Setup}
To systematically benchmark the mathematical reasoning capabilities of existing LLMs, we include a comprehensive evaluation of 43 models, comprising 26 LLMs and 17 MLLMs. These models are principally divided into two categories: generalist models encompassing both API-based commercial LLMs and open-sourced LLMs (large and small scale), and specialized mathematical models. We use the F1 metric for Outcome Judging and Answerable Judging tasks, and the Acc metric for the other two tasks. The list of selected models and details of evaluation setup can be found in Appendix \ref{Evaluation Setup}.% and \ref{Prompt List}.

% -------------------------- Results ---------------------------------

\subsection{Main Results}
\label{main results}
Tables \ref{table_1} and \ref{table_2} illustrate the performance of various models on the \textsc{MathCheck-GSM} and \textsc{MathCheck-GEO}, respectively. The leftmost column represents the average performance across all tasks and all question variants. The middle four columns detail the performance on various mathematical reasoning tasks, while the right four columns display performance across different question variants. Consequently, each model is represented by a 4$\times$4 checklist table, which showcases the model's performance in various dimensions. The details of all checklist tables are further elaborated in Appendix \ref{sec-HeatMapGSM} and \ref{sec-HeatMapGEO}.

% On \textsc{MathCheck-GSM}, GPT-4o achieves the best performance, attaining a remarkable 92.0 on the All score. It also achieved the state of the art in most tasks and robust variants. The GPT-4 model follows closely with 90.9, and had the best result in the answerable judging task. They show that they have excellent mathematical reasoning ability on GSM8k level. In the open source LLMs, the Llama3-70B-Instruct achieved the best score, reaching 84.7 , while showing good performance on various tasks and problem variants. It is worth mentioning that Qwen-72B has poor performance on other tasks except solving, which we suspect is due to its special optimization of solving task. This phenomenon also shows up in all mathematical models.

On \textsc{MathCheck-GSM} (Table~\ref{table_1}), O1-preview and O1-mini exhibit outstanding performance with impressive overall score of 93.2 and 92.7, demonstrates strong effect of extending reasoning thought exploration.
GPT-4o is closely followed with a score of 92.0 and demonstrates top performance on the problem solving task and irrelevant disturbance variants. These results indicate that strong foundational models still possess formidable and robust performance across a variety of mathematical reasoning tasks.
% Surprisingly, we find that O1-mini is only 86.9, it seems extending reasoning length on a small model can't achieve significant improvement.
Among the open-source LLMs, LlaMa-3.1-70B-Instruct achieves the highest score of 90.5 and performs excellently across a range of tasks and problem variants. Its performance has significantly improved compared to LLaMA-3 version and surpasses that of GPT-4o-mini. Besides, Qwen1.5-72B-Chat underperforms in tasks other than problem solving, which we suspect is due to its special optimization of the solving task. This phenomenon is also observed across all math-customized models, which tend to be trained on similar mathematical problems and problem-solving processes, resulting in a relatively narrow scope of reasoning capabilities. 
% In fact, our experiments reveal that smaller SFT models (typically under 20 billion parameters) for mathematical reasoning could impede their ability to follow instructions in other math-related tasks.

% On \textsc{MathCheck-GEO}, GPT-4o also achieves the best performance, attaining 65.3 on the All score. The performance of GPT44-turbo-0409 and GPT4-Vision-Preview is similar, achieve 61.7 and 60.0 respectively. In particular, Claude-3-sonnet-20240229 achieved slightly better results than Claude-3-opus-20240229. In open source MLLMs, solving performance is poor, which indicates that the multi-modal reasoning ability of open source MLLMs still has huge room for improvement.

On \textsc{MathCheck-GEO} (Table~\ref{table_2}), GPT-4o demonstrates the best performance, achieving a top score of 65.3 in the All category. The performance of GPT4-turbo-20240409 and GPT4-Vision-Preview is similar, reaching scores of 61.7 and 60.0, respectively. In particular, the performance of Claude-3-sonnet is slightly superior in visual contexts compared to that of its larger counterpart, Claude-3-opus. 
Among the open-source MLLMs, the large-size MLLMs demonstrate surprisingly strong performance, with Qwen-VL-70B attaining 60.4 over the GPT-4-Vision-Preview.
However, the most of small-size MLLMs exhibited poor performance especially in probelm solving,
% particularly when compared with Answerable and Outcome Judging.
% which exhibits a higher likelihood of successful guessing. 
which suggests that the multi-modal reasoning capabilities of open-source small-size open-source MLLMs still have significant room for improvement.
% meriting further exploration by the research community.

\begin{figure*}[ht!]
    \centering
    \includegraphics[scale=0.175]{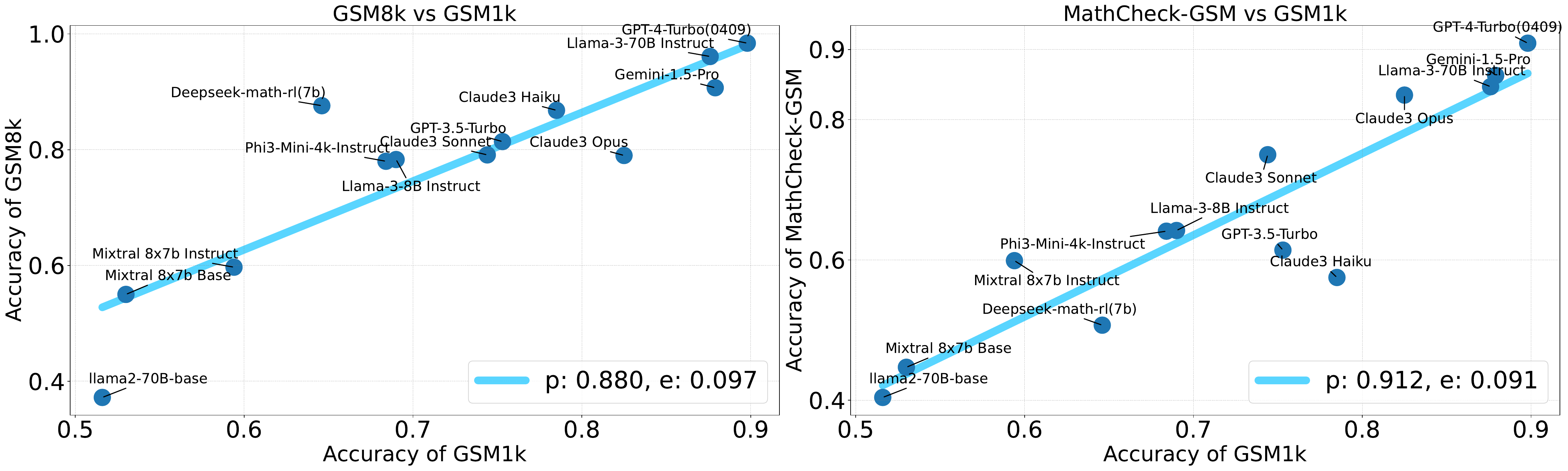}
    \caption{Correlation with GSM1k~\citep{zhang2024careful}, a dataset that reflects real mathematical reasoning ability.  $p$ and $e$ represent the Pearson Correlation Coefficient, and Root Mean Square Error, respectively.}
    \label{fig:gsm1k}
\end{figure*}

% $x$-axis represents the accuracy on GSM1K while the $y$-axis means the accuracy on GSM8k (left) and our MathCheck-GSM (right).

% \subsection{{MathCheck} is Closer Real Reasoning Ability}
\subsection{{MathCheck} Represents Mathematical Intelligence More Linearly}
One desiderata of a good mathematical benchmark is to reflect real mathematical intelligence perfectly.
We follow previous works \citep{zhang2024careful, huang2024compression} to assess ``intelligence" from practical standpoints and use performance on private data \citep{zhang2024careful} and compression efficiency \citep{du2024understanding, huang2024compression} as surrogates to assess the genuine mathematical abilities of models. By examining the correlation between \textsc{MathCheck} and these surrogates, we can verify whether our design effectively reflects mathematical intelligence, and how it compares to traditional benchmarks.
% While reaching a consensus on the precise definition of "intelligence" is challenging, we can follow previous works \citep{zhang2024careful, huang2024compression} to assess "intelligence" from practical standpoints. 
% In this work, we use performance on private data \citep{zhang2024careful} and compression efficiency \citep{du2024understanding, huang2024compression} as surrogates to assess the genuine mathematical abilities of models. 

% Although our \textsc{MathCheck} is a more comprehensive benchmark design, we want to see whether it can truly reflect the mathematical reasoning ability of models.

\textbf{Correlation with Private Data. } Unlike traditional open-sourced benchmarks, private data is less likely to be contaminated or overfitted, making it an appropriate proxy of genuine mathematical intelligence. We adopt GSM1k~\citep{zhang2024careful}, a new private GSM8k-level dataset, to measure the real mathematical reasoning of models. We compare the correlation of model performance between GSM1k and \textsc{MathCheck-GSM}/GSM8k. As shown in Figure ~\ref{fig:gsm1k}, the left part illustrates the correlation between GSM8k and GSM1k. It reveals that most LLMs achieve scores up to 80\% on GSM8k, with scores concentrated in the top half of the graph. However, on GSM1k, the scores are evenly distributed, indicating that some LLMs, such as deepseek-math-7B-RL, have inflated scores on GSM8k. This suggests that the GSM8k score is not a reliable benchmark for assessing the true mathematical reasoning ability of the models. In the right sub-figure, \textsc{MathCheck-GSM} and GSM1k display a good positive correlation, and some models that do not perform well on GSM1k can be detected by \textsc{MathCheck-GSM}. By comparing the Pearson correlation coefficient and the root mean square error, it shows that \textsc{MathCheck} has a higher correlation coefficient with GSM1k, mitigating bias evaluation caused by overfitting and data contamination.

\begin{figure*}[t!]
    \centering
    \includegraphics[scale=0.175]{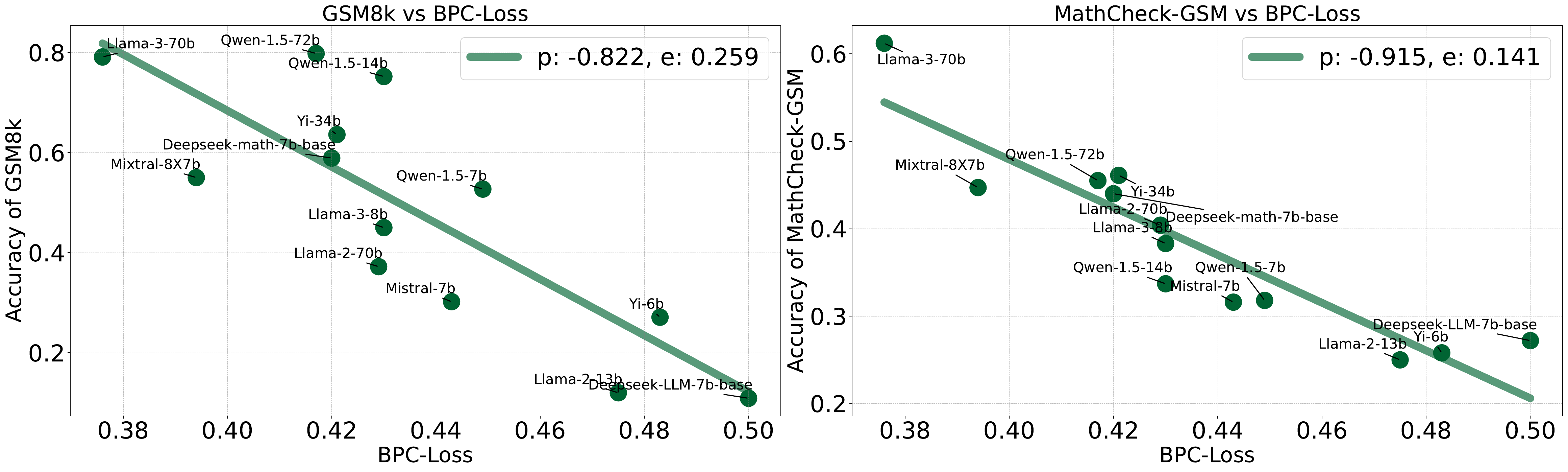}
    \caption{Performance correlation with BPC-loss, which reflects compression efficiency~\citep{huang2024compression}. The lower BPC-loss represents the higher compression efficiency.}
    \label{fig:loss}
\end{figure*}

\textbf{Correlation with Compression Efficiency.}
Compression efficiency has been empirically proven that represent intelligence well \citep{du2024understanding} even linearly \citep{huang2024compression}, well aligned with the belief that compression is closely connected to intelligence \citep{deletang2024language}. Following \cite{huang2024compression}, we use BPC-Loss in Arxiv papers tagged with ``Math" to measure compression efficiency as a surrogate. Figure \ref{fig:loss} shows the correlation between BPC-Loss and GSM8K/MathCheck-GSM. The left sub-figure reveals that a single traditional benchmark like GSM8K cannot adequately reflect genuine mathematical ability, as indicated by the low Pearson correlation coefficient ($p=-0.822$). Many models, such as the Qwen series, deviate significantly from the regression line. In contrast, the right sub-figure displays the correlation with our \textsc{MathCheck-GSM}, demonstrating that \textsc{MathCheck-GSM} exhibits a significantly better correlation with genuine intelligence, with a Pearson correlation coefficient of $p=-0.915$. Our method shows that many models, such as the Qwen series, have scores on our benchmark that align more accurately with their true mathematical abilities. This shows that our design can represent mathematical intelligence more linearly compared to traditional benchmarks.
% By comparing the compression efficiency with math benchmark score, previous work~\citep{huang2024compression} found that the benchmark score is linearly correlates to the compression efficiency, which assures that the compression represents intelligence Linearly. 
% However, in their work, they found that some models deviated from the regression line since data contamination like Qwen-series. 
% In our work, we compare the GSM8k and \textsc{MathCheck-GSM} in correlation with compression efficiency. 
% As shown in Figure.~\ref{fig:loss}, there is also a phenomenon of Qwen series offset when using our GSM8k as math benchmark, and the  pearson correlation coefficient is $p=-0.822$.
% Applying \textsc{MathCheck-GSM}, we can get $p=-0.915$ on pearson correlation coefficient which becoming more linear and Qwen-series no longer deviate from the regression line. These findings indicates \textsc{MathCheck-GSM} is a better proxy for intelligent benchmarks.

% \begin{wrapfigure}{r}{0pt}
% \centering
% \includegraphics[width=0.2\textwidth]{C:/Users/ASUS/Pictures/5.png}   
% \caption{\footnotesize maka}
% \end{wrapfigure}

\begin{figure*}[t!]
    \centering
    \includegraphics[scale=0.395]{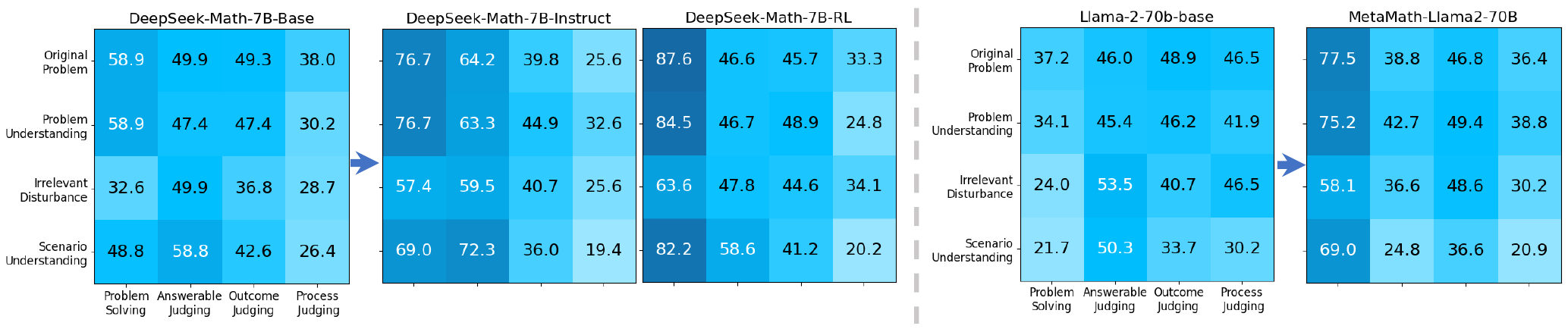}
    \caption{Behavior of mathematical models trained on 
massive solving data.}
    \label{fig:sft}
\end{figure*}

\section{Behavior Analysis}
\textsc{MathCheck} contains multi-dimensional information for evaluation, therefore we can observe the behaviors of the models on it to help analyze the models.

\textbf{Behavior of Math Models. }
% Recently, some works trained on massive solving data to improve the math solving performance, then come to the conclusion that the reasoning ability is greatly improved. 
% Recently, some works claim the reasoning ability is greatly improved by trained on massive solving data.
Recently, some works claim that math reasoning ability is greatly improved by training on massive amounts of math solving data, but this may not necessarily be true. To validate whether their mathematical reasoning ability really improves, we examine the behaviors of the math models and their base models on \textsc{Mathcheck}. As shown in Figure~\ref{fig:sft}, compared with the base model, the performance of DeepSeek-Math-7B-Instruct/RL on solving units is greatly improved. However, the performance improvement on other units is limited, or even downward. The same phenomenon can be observed on MetaMath. It implies that training solely on massive solving data is not the right direction to improve mathematical reasoning ability. Instead, training models with high-quality and diverse mathematical data, beyond just solving, should be considered.

\textbf{Reasoning Consistency. }
We analyze the reasoning consistency of generalist models across each unit in \textsc{MathCheck}, and the detailed results are shown in Appendix \ref{sec-HeatMapGSM} and \ref{sec-HeatMapGEO}. We can see most of them show good reasoning consistency since they achieve similar scores on each unit, such as GPT series, Llama-3 series and Mixtral series on \textsc{MathCheck-GSM} and GPT series on \textsc{MathCheck-Geo}. This is an interesting finding as it substantiates our assertion: \textit{a model that really understands a problem can robustly work well on multiple related tasks}. Meanwhile, we also find that some models perform reasoning inconsistently. For example, Qwen1.5-72B-chat, Claude-3-Haiku and Phi-3-Medium show excellent performance on the solving task but much worse in other units of \textsc{MathCheck-GSM}. On \textsc{MathCheck-GSM}, Internet-VL achieves a high score of 40.0 on the original problem solving but decreases considerably when the problem switches to other robustness variants. These abnormal inconsistency behaviors of generalist models are highly similar to those mathematical models, revealing that they may conduct excessive decoration on original benchmarks. 
%which can be divided into two categories. One is that the score on solving task is lower, such as CogVLM2 on \textsc{MathCheck-GEO}. 
%It is a normal phenomenon since the solving task is the most difficult among these four tasks. Though the model cannot solve the problem well, it still exhibits some mathematical ability on math judging task. The emergence of problem-solving abilities requires either an expansion of their parameters or data. Another inconsistency is an obvious higher scores on solving task or original problems. On \textsc{MathCheck-GSM}, Qwen1.5-72B-chat, Claude-3-Haiku and Phi-3-Medium show excellent performance on the solving task but much worse in other units.
% This phenomenon indicates its over-fitting in the solving task, which cannot correctly reflect its real mathematical reasoning ability.
%On \textsc{MathCheck-GSM}, Internet-VL achieves a high score of 40.0 on the original problem solving but decreasing much when the problem becomes another robustness variants. These abnormal inconsistency behaviors are highly similar to the math models, which reveals they conduct excessive decoration on original benchmark.

% \begin{wrapfigure}{r}{0.45\textwidth}%靠文字内容的右侧,改width和r/l,2cm等
%     \centering
%     \includegraphics[width=0.45\textwidth]{Styles/gsm-prompting.pdf}
%     \caption{Different prompting strategies on \textsc{MathCheck-GSM}}
%     % \vspace{-1cm} 
%     \label{gsm_prompting}
% \end{wrapfigure}

\begin{figure}[t]
\begin{minipage}{0.58\textwidth}
\centering
% \hspace{-5mm}
\includegraphics[width=1\linewidth]{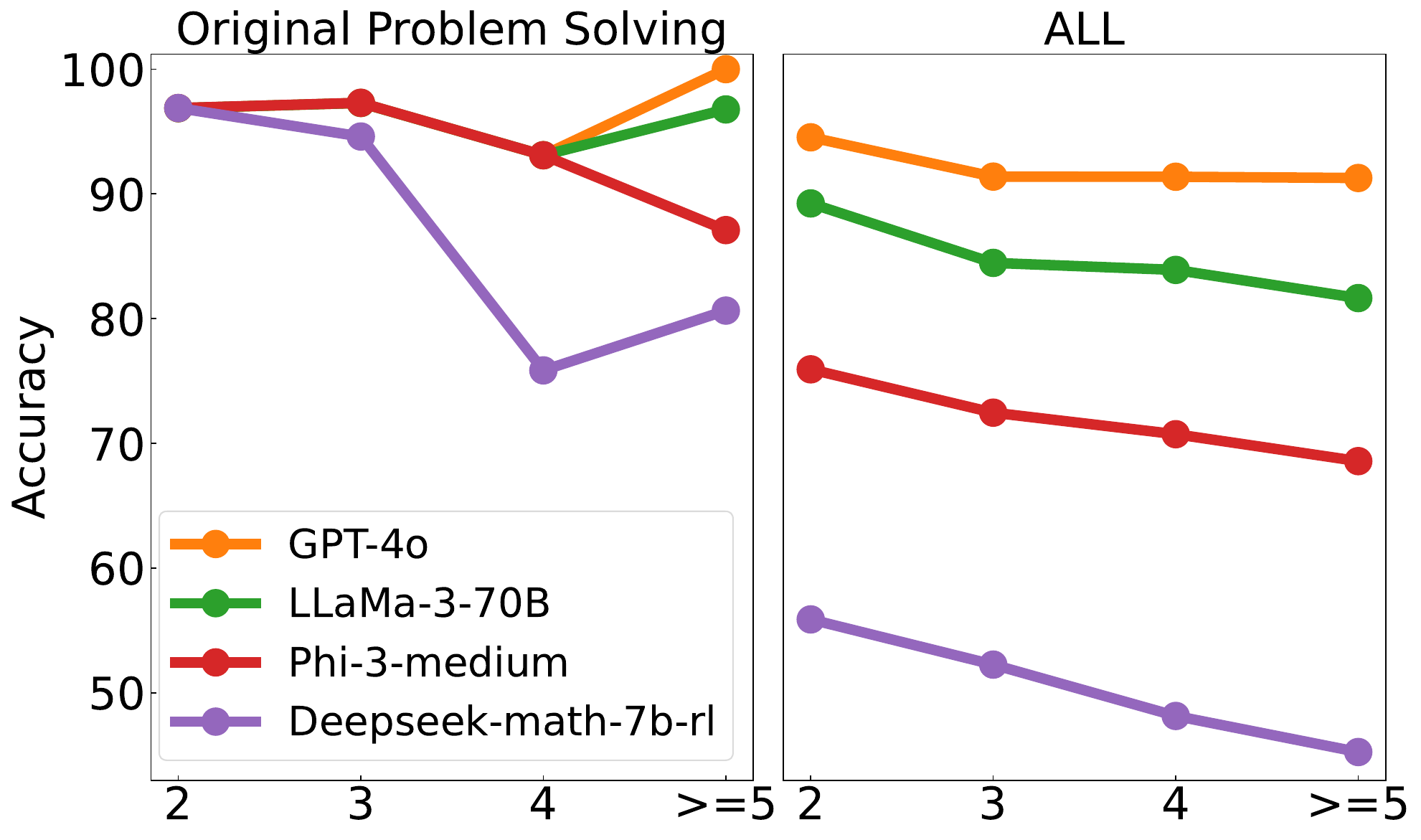}
% \vspace{-5mm}
 \caption{Performance on different complexity levels (i.e., reasoning steps) of \textsc{MathCheck-GSM}.}
 \label{fig:complexity levels}
 \end{minipage} 
 \hfill
 % \hspace{2mm}
 \begin{minipage}{0.392\textwidth}
 \centering
 % \vspace{5mm}
\includegraphics[width=1\linewidth]{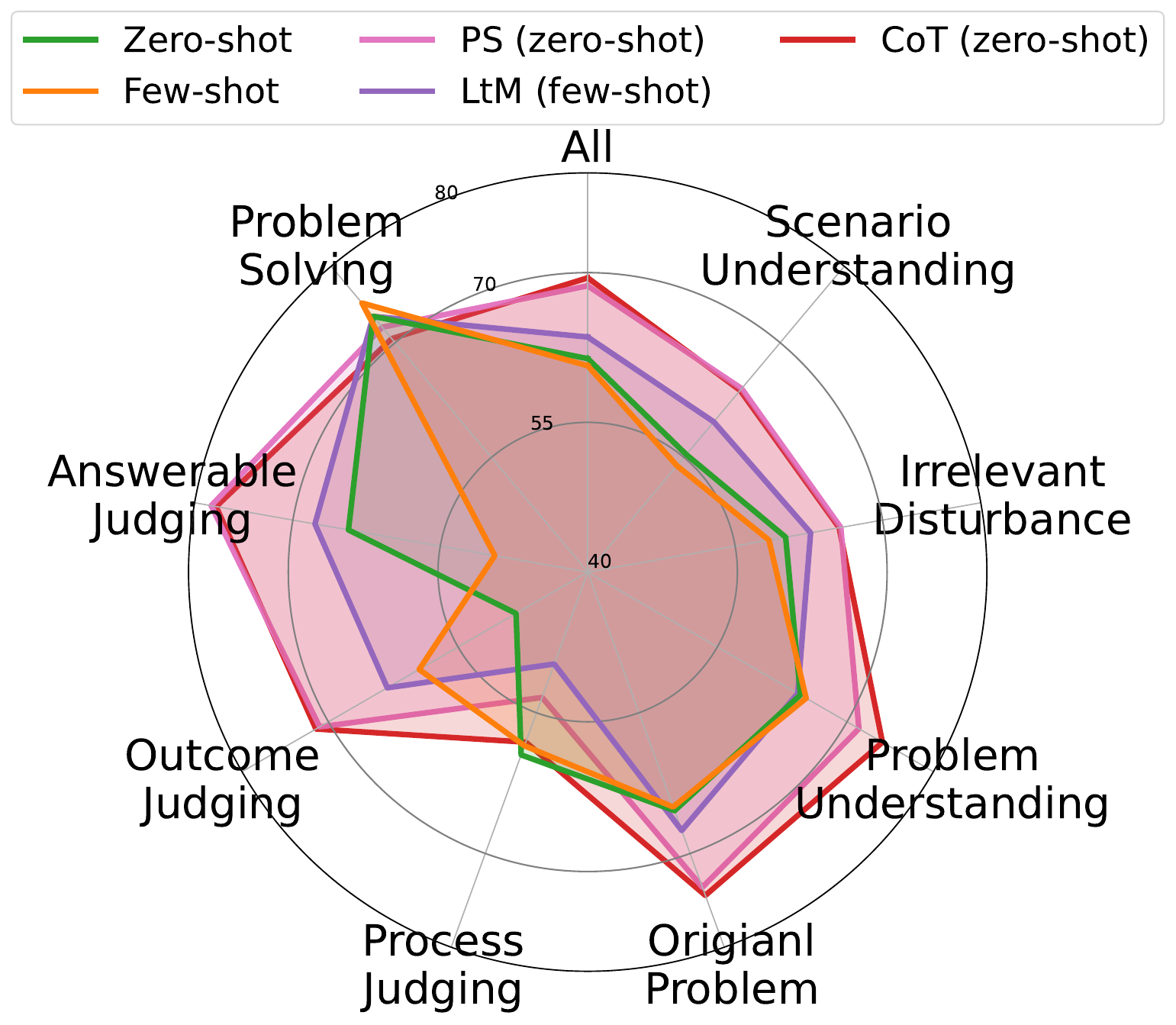}
% \vspace{-3mm}
 \caption{Different prompting technologies on \textsc{MathCheck-GSM}.}
 \label{fig:prompting technologies}
 \end{minipage}
 % \vspace{-2mm}
\end{figure}

\textbf{Behavior on Different Complexity Levels.} 
We categorize the complexity of problems based on the number of reasoning steps of the original problems, and select representative models of varying sizes for evaluation, as depicted in Figure \ref{fig:complexity levels}. We can observe that the models' accuracy on the original problem solving fluctuates and does not show an obvious downward trend as the problems are more difficult. While the score ''All'' shows a steady downward trend, it implies that \textsc{MathCheck} better demonstrates the reasoning skills and capabilities required when problems become difficult.

% The results indicate that most problems of varying complexity associated with the seed questions (GSM8K) are effectively resolved. However, the challenges posed by the \textsc{MathCheck-GSM} dataset are notably more difficult. Additionally, as the complexity increased, the difficulty levels rose steadily. This highlights that our \textsc{MathCheck} represents a more challenging and robust multi-task benchmark.

\textbf{Behavior on Different Prompting Technologies.}
We evaluate five prompting techniques including Zero-shot, Few-shot~\citep{brown2020language}, CoT~\citep{wei2022chain}, Least to Most prompting~\citep{zhou2022least}, and Plan-and-Solve prompting~\citep{wang2023plan}. The results of GPT-3.5-Turbo on \textsc{MathCheck-GSM} are illustrated in Figure \ref{fig:prompting technologies}. Overall,  Chain of Thought (CoT) and Plan-and-Solve (PS) in the zero-shot setting demonstrate superior performance, though this is not consistently the case across all tasks and settings. In contrast, the Few-shot prompt generally yields worse results than the Zero-shot prompt. % except those in Problem Solving tasks. 
% Our analysis of the model prediction results indicates that the few-shot setting tends to diminish and abbreviate the reasoning steps. 
Through detailed analyses, we find that the math reasoning generalization of LLMs is sensitive to Few-shot samples, which inspires us that Zero-shot with advanced prompt techniques (e.g., CoT or PS) may be a better choice in mathematical reasoning tasks.

% Plan-and-Solve (PS) prompt also does not achieve satisfactory results, potentially due to its problem-decomposition method being less effective for tasks beyond Problem Solving.

\section{\textsc{MathCheck} applied to other  reasoning tasks}
Our proposed benchmark paradigm \textsc{MathCheck} can be adapted to other reasoning tasks beyond mathematical problems. Similarly, it can help other reasoning tasks with more comprehensive evaluations and conduct detailed analysis through model behaviors. We attempt the migration of the \textsc{MathCheck} paradigm in both commonsense reasoning and code generation.

%We firstly adapt 
Commonsense reasoning requires LLMs to apply parametric knowledge to reason and solve problems. In this paper, we choose the date understanding task in Big-bench~\citep{srivastava2023beyond} as test-bed since it is wildly used to measure commonsense reasoning ability~\citep{wei2022chain}. Appendix~\ref{sec:date} shows the case of applying \textsc{MathCheck} to date understanding. Similar to mathematical reasoning, date understanding is a numerical reasoning task, where it can easily utilize variants of each unit in \textsc{MathCheck}. With \textsc{MathCheck}, a simple raw data of date understanding
% (Yesterday‘s date was 4/30/2021. What is the date tomorrow in MM/DD/YYYY?) 
have various corresponding test cases to examine the reasoning robustness and task generalization, helping us better evaluate model's understanding of dates and avoiding hallucination. This case demonstrates the high adaptability of \textsc{MathCheck} to commonsense reasoning task especially numerical reasoning. 

In addition to commonsense reasoning task, we would like to show the possibility of transforming \textsc{MathCheck} in some real-world reasoning tasks such as code generation. Appendix~\ref{sec:code} demonstrates a case of applying \textsc{MathCheck} to code generation. Unlike numerical reasoning tasks, the adaptation of code generation needs to consider task relevance. It shows that \textsc{MathCheck} has potential for real-world tasks such as agents and robotics application. For such real-world tasks, it is important to have multiple variants to examine reasoning robustness and task generalization, which directly reflects the diversity of environment and user requirements. Meanwhile, we encourage researchers to design more variants towards specific reasoning tasks beyond \textsc{MathCheck} and promote research on general reasoning benchmark paradigm.

%--------------------- Related work-------------------------------

\section{Related Work}
\textbf{Benchmarks of Textual Mathematical Reasoning. }
Numerous benchmarks have been proposed to evaluate the mathematical reasoning capabilities of AI models, including~\citep{amini2019mathqa,cobbe2021training,frieder2024mathematical}. Some datasets, such as the elementary-level GSM8k~\citep{cobbe2021training}. Consequently, more challenging datasets have been introduced, including those at the high-school level~\citep{hendrycks2021measuring}, university level ~\citep{sawada2023arb,zheng2021minif2f} and olympic level~\citep{huang2024olympicarena}.
Additionally, to provide a more comprehensive evaluation of  mathematical reasoning abilities, numerous benchmarks have been developed that measure the robustness of mathematical reasoning~\citep{li2024gsm}, including semantic perturbations~\citep{wang2023large,zhou2024mathattack}, reverse problem-solving~\citep{yu2023metamath,berglund2023reversal}, and irrelevant distractions~\citep{shi2023large,li2023you}.
Above benchmarks paradigm focus on a single task (typically problem-solving) and can not comprehensively reflect models' reasoning ability at a given level. 
Therefore, \textsc{MathCheck} tries to go for better reasoning benchmark paradigm.
% explores the task generalizability of mathematical reasoning. It also covers different types of robustness assessment across various tasks, providing a comprehensive evaluation of mathematical reasoning capabilities.

\textbf{Benchmarks of Visual Mathematical Reasoning. }Recently, multi-modal large language models have demonstrated outstanding capabilities in visual-language reasoning tasks~\citep{allaway2022penguins,chen2023pali,yang2023dawn,team2023gemini}. Several benchmarks~\citep{lin2014microsoft,antol2015vqa,hudson2019gqa,marino2019ok,mobasher2022parsvqa} have been introduced to assess the visual reasoning capabilities of multi-modal large language models across various modalities including abstract scenes, geometric diagrams, graphics, and charts~\citep{lu2021inter,chen2021geoqa,chen2022unigeo,masry2022chartqa,kazemi2023geomverse}. To integrate various math visual reasoning tasks, \cite{lu2023mathvista} propose MathVista by collecting multiple datasets. 
Currently, research on visual mathematical reasoning remains at a preliminary stage, 
% particularly with regard to the problem-solving capabilities of open-source MLLMs, which exhibit rudimentary problem-solving performance (see our experimental results in Section~\ref{main results}).
% Our benchmark provides a comprehensive assessment of this field, focusing on task generalizability and problem robustness within visual contexts.
Our benchmark focuses on the universality of tasks and robustness of question formulation within visual contexts, offering a comprehensive evaluation and testing platform for the research community.

%--------------------- Conclusion-------------------------------

\section{Conclusion}
In this paper, we argue that if a model really understands a problem, it should be able to successfully solve various tasks and variations of that problem. Based on this insight, we introduce \textbf{\textsc{MathCheck}}, a well-designed checklist for testing task generalization and reasoning robustness. To this end, we also propose an automatic tool for efficiently generating checklist for most of math reasoning datasets. Our proposed \textsc{MathCheck} allows the research community to clearly observe model performance across different dimensions, yielding more comprehensive and objective evaluation results. Using \textsc{MathCheck}, we develop \textbf{\textsc{MathCheck-GSM}} for textual reasoning and \textbf{\textsc{MathCheck-GEO}} for multi-modal reasoning. In the experiments, we evaluate massive (M)LLMs and conduct detailed analysis of model behaviors on \textsc{MathCheck}. Subsequently, we reveal that
the evaluation on \textsc{MathCheck} is much closer to the true reasoning abilities than previous benchmark datasets (e.g., GSM8K and Geometry3K). Finally, we also show the potential of applying \textsc{MathCheck} paradigm to other reasoning tasks such as commonsense reasoning and code generation. We hope our practice and observation can constitute a significant stride towards better reasoning benchmark paradigm.

\bibliography{iclr2025_conference}
\bibliographystyle{iclr2025_conference}

% \section{Limitation}\label{limit}
% In this section, we discuss the limitations of our work. First of all, 
% % \textsc{MathCheck} is an automatic method for generating multiple math tasks data and  diverse problem variants. However,  we did not adopt more mathematical task types and robust variants considering the high quality of the generated data. 
% we have evaluated 31 popular models but not all  mainstream models on \textsc{MathCheck} owing to the limited computational resources and API-cost. To this end, we release the dataset with evaluation code to expediently evaluate new models (e.g., Reka, Grok) .
% Second, we focus on the evaluation of real mathematical reasoning ability of large models in this work, and explore the impact of different methods on reasoning ability from the perspective of training data and prompt strategies; nevertheless, we do not propose an effective and innovative method to improve mathematical reasoning ability, which will be carried out as our future work.

% \newpage
\clearpage
\appendix

% Reset depth to add sections and subsections to ToC
\addtocontents{toc}{\protect\setcounter{tocdepth}{3}}

% Setting colorlinks=black just for the table of contents
% \hypersetup{linkcolor=black}
% \hypersetup{
% 	colorlinks=false,
% 	linkcolor=black,
% 	filecolor=black,      
% 	urlcolor=black,
% 	citecolor=black,
% }
\hypersetup{
linkcolor=black,
pdfborder={0 0 0} % 设置边框宽度为0
}
\renewcommand*\contentsname{Appendix}
\tableofcontents % Lists only the appendix sections and subsections

\hypersetup{
% linkcolor=red,
pdfborder={1 1 1} % 设置边框宽度为0
}

\newpage
\section{Heatmap of \textsc{MathCheck-GSM}}
\label{sec-HeatMapGSM}
% We provide the performance of all checklists on the \textsc{MathCheck-GSM} dataset as shown in Figure \ref{fig:heatmap_gsm}.

\begin{figure*}[h!]
    \centering
    \includegraphics[scale=0.65]{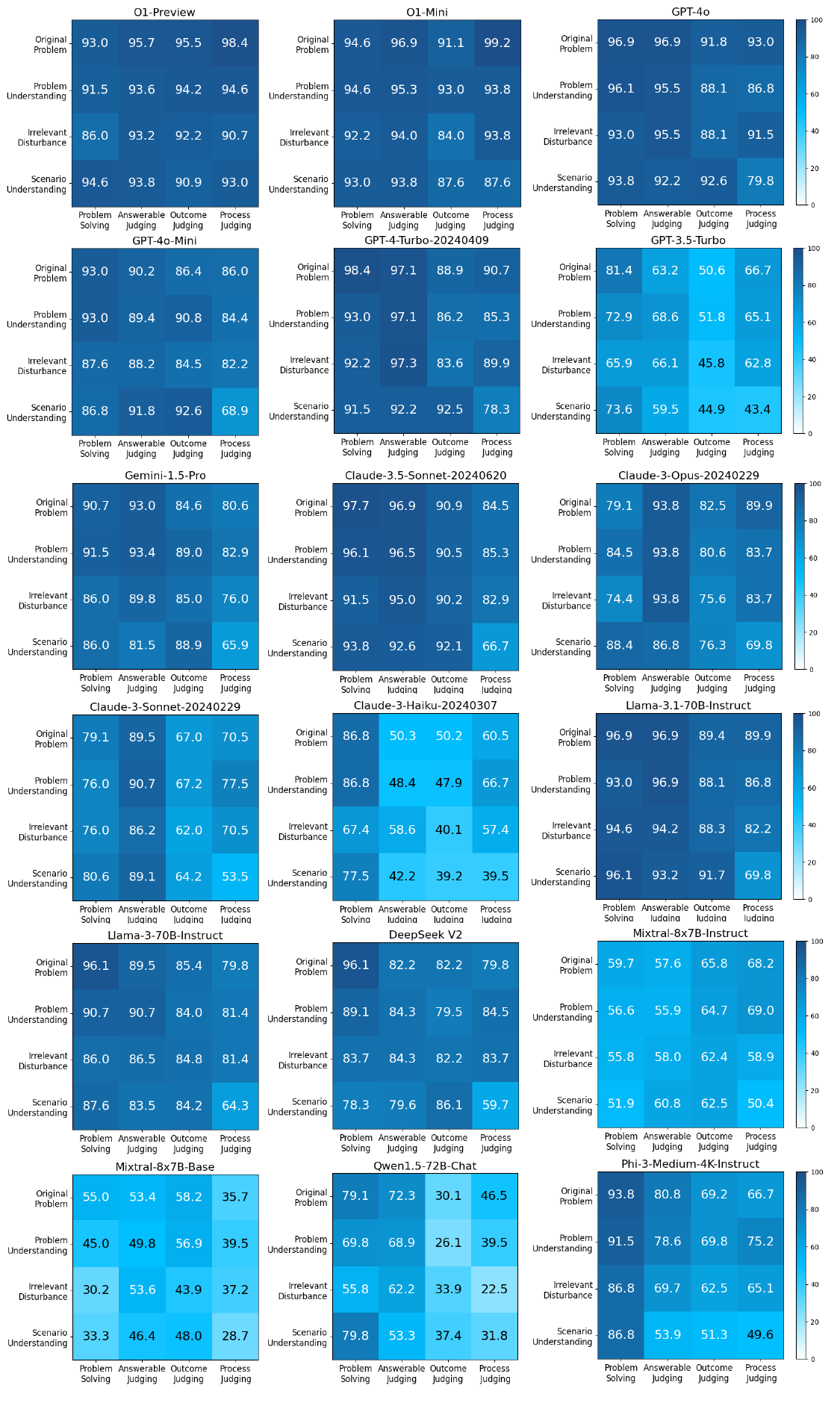}
    \caption{Visualized heatmap of \textsc{MathCheck-GSM} - Part 1.}
    \label{fig:heatmap_gsm_1}
\end{figure*}

\newpage

\begin{figure*}[h!]
    \centering
    \includegraphics[scale=0.65]{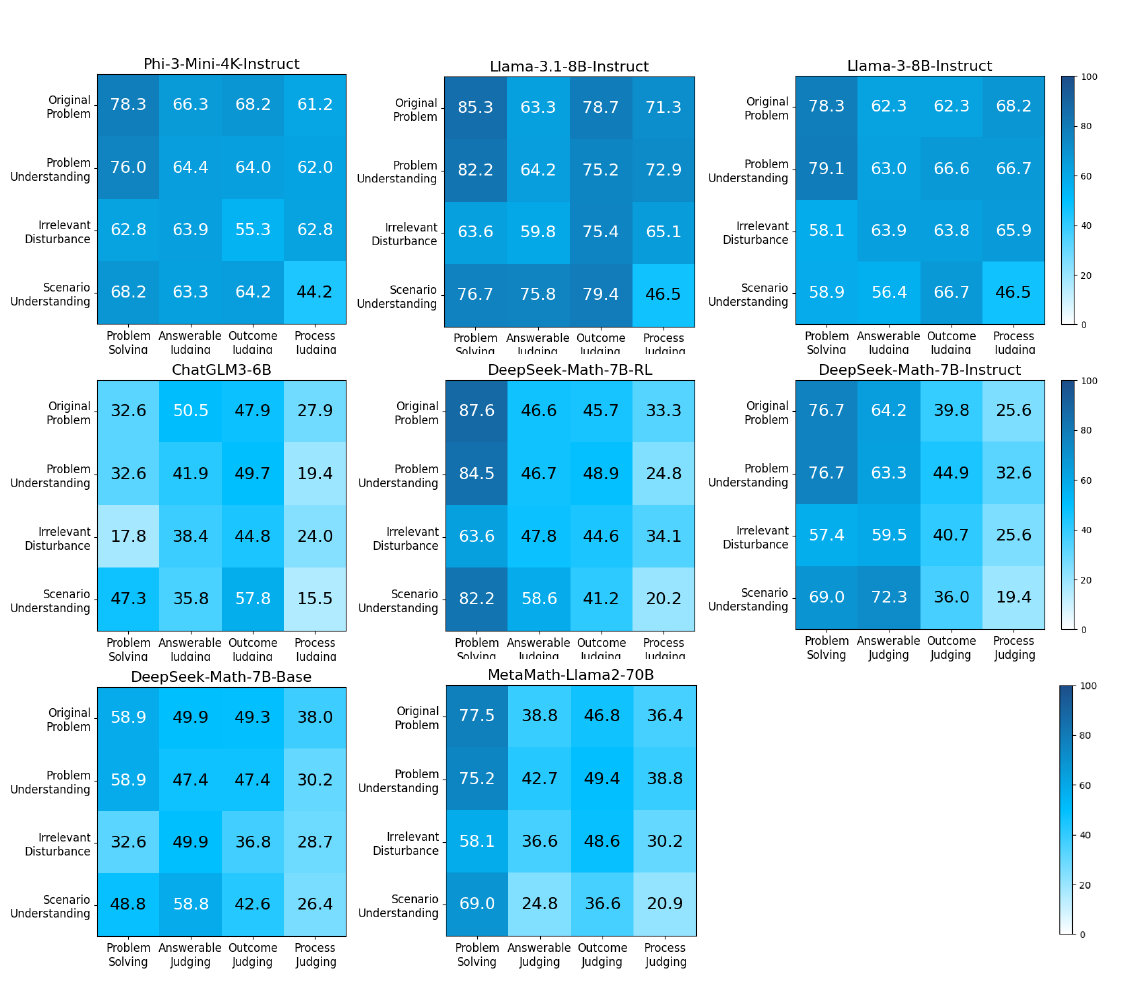}
    \caption{Visualized heatmap of \textsc{MathCheck-GSM} - Part 2.}
    \label{fig:heatmap_gsm_2}
\end{figure*}

\newpage

\section{Heatmap of \textsc{MathCheck-GEO}}
\label{sec-HeatMapGEO}
% We provide the performance of all checklists on the \textsc{MathCheck-GEO} dataset as shown in Figure \ref{fig:heatmap_geo}.

\begin{figure*}[ht!]
    \centering
    \includegraphics[scale=0.65]{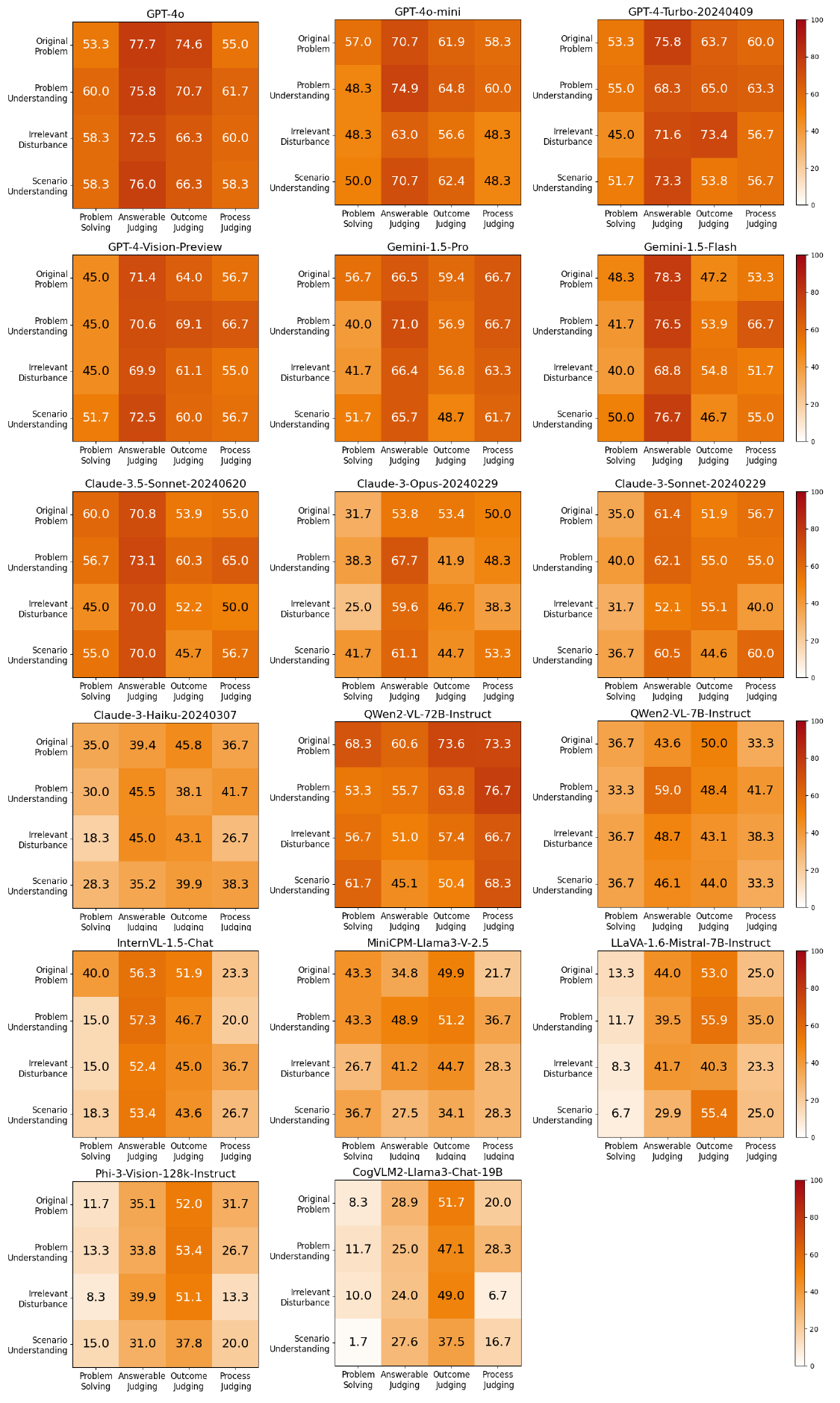}
    \caption{The visualized heatmap of \textsc{MathCheck-GEO}.}
    \label{fig:heatmap_geo}
\end{figure*}
 
\clearpage

\section{Data Statistics and Quality}\label{datasta}
\subsection{Overview of Data}
Table~\ref{gsm_ov} and Table~\ref{geo_ov} show the data statistics of \textsc{MathCheck-GSM} and \textsc{MathCheck-Geo}. Table~\ref{group_ov} shows the data statistics of each group in \textsc{MathCheck-GSM} and \textsc{MathCheck-Geo}. In each group, since answerable judging and outcome judging are binary-classification tasks, we try our best to include two different labels in these units for fair evaluation.

\begin{table*}[ht!]
\centering
\caption{Data statistics of \textsc{MathCheck-GSM}}
% \scalebox{1}{
\begin{tabular}{lrrrr}
\toprule
 & \textbf{\makecell[r]{Problem\\Solving}}
 & \textbf{\makecell[r]{Answerable\\Judging}}
 & \textbf{\makecell[r]{Outcome\\Judging}}
 & \textbf{\makecell[r]{Process\\Judging}} \\
\midrule
\textbf{Original Problem}       & 129 & 258 & 258 & 129 \\
\textbf{Problem Understanding}  & 129 & 258 & 258 & 129 \\
\textbf{Irrelevant Disturbance} & 129 & 258 & 258 & 129 \\
\textbf{Scenario Understanding} & 129 & 258 & 258 & 129 \\
\bottomrule
\end{tabular}
\label{gsm_ov}
\end{table*}

\begin{table*}[ht!]
\centering
\caption{Data statistics of \textsc{MathCheck-GEO}}
\scalebox{1}{
\begin{tabular}{lrrrr}
\toprule
 & \textbf{\makecell[r]{Problem\\Solving}} %Problem\\
 & \textbf{\makecell[r]{Answerable\\Judging}}
 & \textbf{\makecell[r]{Outcome\\Judging}}
 & \textbf{\makecell[r]{Process\\Judging}} \\
\midrule
\textbf{Original Problem}       & 60 & 120 & 120 & 60 \\
\textbf{Problem Understanding}  & 60 & 120 & 120 & 60 \\
\textbf{Irrelevant Disturbance} & 60 & 120 & 120 & 60 \\
\textbf{Scenario Understanding} & 60 & 120 & 120 & 60 \\
\bottomrule
\end{tabular}}
\label{geo_ov}
\end{table*}

\begin{table*}[ht!]
\centering
\caption{Data statistics of each group in \textsc{MathCheck-GSM} and \textsc{MathCheck-GEO}}
\scalebox{1}{
\begin{tabular}{lrrrr}
\toprule
 & \textbf{\makecell[r]{Problem\\Solving}}
 & \textbf{\makecell[r]{Answerable\\Judging}}
 & \textbf{\makecell[r]{Outcome\\Judging}}
 & \textbf{\makecell[r]{Process\\Judging}} \\
\midrule
\textbf{Original Problem}       & 1 & 2 & 2 & 1 \\
\textbf{Problem Understanding}  & 1 & 2 & 2 & 1 \\
\textbf{Irrelevant Disturbance} & 1 & 2 & 2 & 1 \\
\textbf{Scenario Understanding} & 1 & 2 & 2 & 1 \\
\bottomrule
\end{tabular}}
\label{group_ov}
\end{table*}
\newpage

\subsection{Effectiveness of GPT-4-turbo Rewriting}\label{secPR}
In the process of human evaluation, we conduct statistics on the pass rate of \textsc{MathCheck-GSM} rewritten by GPT4-turbo, as shown in Table~\ref{passrate}. It can be seen that the rewriting pass rate is high, which reflects the effectiveness of our generation method. The success rate of Problem Understanding and Scenario Understanding is higher than 90\%. There is a pass rate of 86.82\% in the Irrelevant Disturbance and 81.40\% in Wrong Step Rewriting. It provides references when we use \textsc{MathCheck} generation.

\begin{table*}[ht!]
\centering
\caption{Pass rate (\%) checked by human annotators for the data generated by GPT4-turbo.}
\scalebox{0.89}{
\begin{tabular}{lrrrrr}
\toprule
\textbf{\makecell[l]{Rewriting\\Type}} & 
\textbf{\makecell[r]{Problem\\Understanding}} & 
\textbf{\makecell[r]{Irrelevant\\Disturbance}} & 
\textbf{\makecell[r]{Scenario\\Understanding}} & 
\textbf{\makecell[r]{Unanswerable\\Question Rewriting}} & 
\textbf{\makecell[r]{Wrong Step\\Rewriting}}  \\
\midrule
\textbf{\makecell[l]{Human\\Pass Rate}}& 93.02 & 86.82 & 91.47 & 85.38 & 81.40 \\
\bottomrule
\end{tabular}}
\label{passrate}
\end{table*}

\subsection{Discussion of data bias generated by GPT}\label{bias}
While we acknowledge there are possible self-bias in LLM-rewritten questions, we assert that this bias is acceptable and does not undermine the conclusions or rationality of \textsc{MathCheck}. This is supported by considerations across several dimensions.

\textbf{Motivations. }The motivation behind \textsc{MathCheck} is to establish a paradigm that mitigates benchmark hacking in the evaluation of mathematical reasoning, thereby revealing the genuine mathematical reasoning abilities of language models more comprehensively.
Rewriting is an integral part of the \textsc{MathCheck} pipeline, which can naturally be performed by either humans or LLMs. While we acknowledge that involving experts in the rewriting process might be the fairest approach, the scalability of this method is a significant concern, as noted in several of today’s LLM benchmarks, such as Arena Hard~\citep{li2024crowdsourced} and MT-Bench~\citep{zheng2023judging}, due to the high associated costs. To enhance scalability and practicality, we opted to use LLMs as the rewriters.
Given that GPT-4 is widely recognized as the most advanced model accessible to the public, we believe that choosing GPT-4 as the rewriter is the closest approximation to the quality of expert human rewriting.

\textbf{Human-Checked Questions. }
In fact, for the data construction which the LLM participates in, we mainly utilize the powerful rewriting ability of LLMs to edit the seed math problem instead of generating a new one from scratch. Moreover, we manually check the generated text to avoid some unnatural generated text.

%\textbf{Experimental Results Comparing Non-GPT-Rewritten and GPT-Rewritten Questions. }
\textbf{Experimental Results and Analysis. } On one hand, although the data are generated by GPT-4-Turbo in our experiments, they do not bring extra benefits to GPT-Family models to make them obviously outperform others. As shown in Table~\ref{table_1}, the performance of Claude-3.5-sonnet is similar with GPT-4-Turbo, and even much better than GPT-4o-mini, which follows the commonsense on these LLMs. 
On the other hand, we compare the experimental results on Non-GPT-Rewritten and GPT-Rewritten Questions. In some data constructions where the LLM is not involved, GPT4-family exhibits the same performance ranking as the score ''All''. Specifically, 
% there are three units belongs to Non-GPT-Rewiritten Questions: Original Problem\&Outcome Judging(\textbf{OP-OJ}), Scenario Understanding\&Problem Solving(\textbf{SU-PS}) and Scenario Understanding\&Outcome Judging(\textbf{SU-OJ}).Because the data of Scenario Understanding and Outcome Judging are generated based on the rules.
the samples in Original Problem\&Outcome Judging (\textbf{OP-OJ}) belong to Non-GPT-Rewiritten Questions, which are generated based on the rules.
Table~\ref{table:bias} shows that the performance ranking on non-LLM-generated data is close to the score ''All'' , where GPT-series continues to perform better than other advanced models. All of these results verify that the possible bias to  GPT models is acceptable in our \textsc{MathCheck}.

\begin{table*}[ht!]
\centering
\caption{Model performance on Non-GPT-Rewiritten Questions of \textsc{MathCheck-GSM}}
\begin{tabular}{lllll}
% \hline
% Models                 & ALL  & OP-OJ         & SU-PS         & SU-OJ         \\ \hline
% GPT-4o                 & 92.0 & \textbf{91.8} & \textbf{93.8} & \textbf{92.6} \\
% GPT-4-Turbo-20240409   & 90.9 & \textbf{88.9} & \textbf{91.5} & \textbf{92.5} \\
% Gemini-1.5-Pro         & 86.3 & 84.6          & 86.0          & 88.9          \\
% Claude-3-Opus-20240229 & 83.5 & 82.5          & 88.4          & 76.3          \\
% Llama-3-70B-Instruct   & 84.7 & 85.4          & 87.6          & 84.2          \\ \hline

\hline
Models                 & All  & OP-OJ                 \\ \hline
GPT-4o                 & 92.0 & \textbf{91.8}  \\
GPT-4-Turbo-20240409   & 90.9 & \textbf{88.9}  \\
Gemini-1.5-Pro         & 86.3 & 84.6                    \\
Claude-3-Opus-20240229 & 83.5 & 82.5                    \\
Llama-3-70B-Instruct   & 84.7 & 85.4                    \\ \hline

\end{tabular}
\label{table:bias}
\end{table*}

%\textbf{Experimental Results Comparing Non-GPT-Family SOTA Models and GPT-Family Models. }  
%In order to verify that (1) a model (Claude-3.5-sonnet) with comparable performance to GPT-4o would not perform obviously poorer due to bias (2) a slightly weaker GPT model (GPT-4o-mini) will not be better than stronger models (Claude-3.5-sonnet) form other family due to bias. We evaluate the performance of Claude-3.5-sonnet and GPT-4o-mini on the \textsc{MathCheck-GSM}. As shown in Table~\ref{table_1}, Claude-3.5-sonnet achieves comparable performance with GPT-4o/4turbo. GPT-4o-mini performs similarly with the Gemini-1.5-Pro but can't outperform the Claude-3.5-sonnet.

\section{Evaluation Setup}
\label{Evaluation Setup}
We conduct evaluations of multiple representative generalist and mathematical models on our \textsc{MathCheck} benchmark. 
For \textsc{MathCheck-GSM}, the evaluation models encompass: (a) Generalist models, including proprietary models such as O1-Preview~\citep{O1-Preview}, O1-Mini~\citep{O1-Mini}, GPT-4o~\citep{GPT-4o}, GPT-4o-mini~\citep{GPT-4o-mini}, GPT-4-Turbo~\citep{achiam2023gpt}, GPT-3.5-Turbo~\citep{turbo}, Gemini-1.5-Pro~\citep{team2023gemini}, Claude-3~\citep{claude3}, Claude-3.5-Sonnet~\cite{claude3-5}, Llama-3\footnote{\url{https://ai.meta.com/blog/meta-llama-3}}, Llama-3.1\footnote{\url{https://ai.meta.com/blog/meta-llama-3-1}}, DeepSeek V2~\citep{shao2024deepseekmath}, Mixtral 8 x 7B~\citep{jiang2024mixtral}, Qwen1.5~\citep{bai2023qwen}, Phi-3~\citep{abdin2024phi}, and ChatGLM3~\citep{du2022glm}; (b) Mathematical models, including DeepSeek-Math~\citep{shao2024deepseekmath} and MetaMath~\citep{yu2023metamath}. 
For \textsc{MathCheck-GEO}, we conduct evaluations on generalist models: (a) proprietary models such as GPT-4o~\citep{GPT-4o}, GPT-4o-mini~\citep{GPT-4o-mini}, GPT-4-Turbo~\citep{achiam2023gpt}, GPT-4-vision~\citep{GPT4V}, Gemini-1.5-Pro~\citep{team2023gemini}, Claude-3.5-Sonnet~\cite{claude3-5} and Claude-3~\citep{claude3}; (b) open-source models including Qwen2-VL~\citep{wang2024qwen2}, InternVL-1.5~\citep{chen2023internvl}, Phi-3-Vision~\citep{abdin2024phi}, LLaVA-1.6-Mistral-7B-Instruct~\citep{liu2024visual}, MiniCPM-Llama3-V-2.5~\citep{yao2024minicpmv} and CogVLM2-Llama3~\citep{wang2023cogvlm}.

For Problem Solving and Process Judging tasks, we employ accuracy as the evaluation measure. For Outcome Judging and Answerable Judging tasks, we utilize Macro-F1 as the metric. We employ a zero-shot setting for generalist models and a few-shot setting (two-shot) for base models and mathematical models to enhance their ability to follow specific instructions and tasks. All the prompts used for evaluating (M)LLMs are provided in Appendix~\ref{secEvaPrompt}.

For all the close-resourced models, we utilize the default hyper-parameters, setting the temperature to 0 and the max tokens to 1,024. Similarly, for all open-source models, the parameters are uniformly configured as follows: $do\_sample$ is set to False, $max\_gen\_len$ is set to 512, and the temperature is set to 0.1.

\clearpage

\section{\textsc{MathCheck} Applied to Other Reasoning Tasks}

\label{mathcheck_other}
\begin{figure*}[hp]
    \centering
    \includegraphics[scale=0.47]{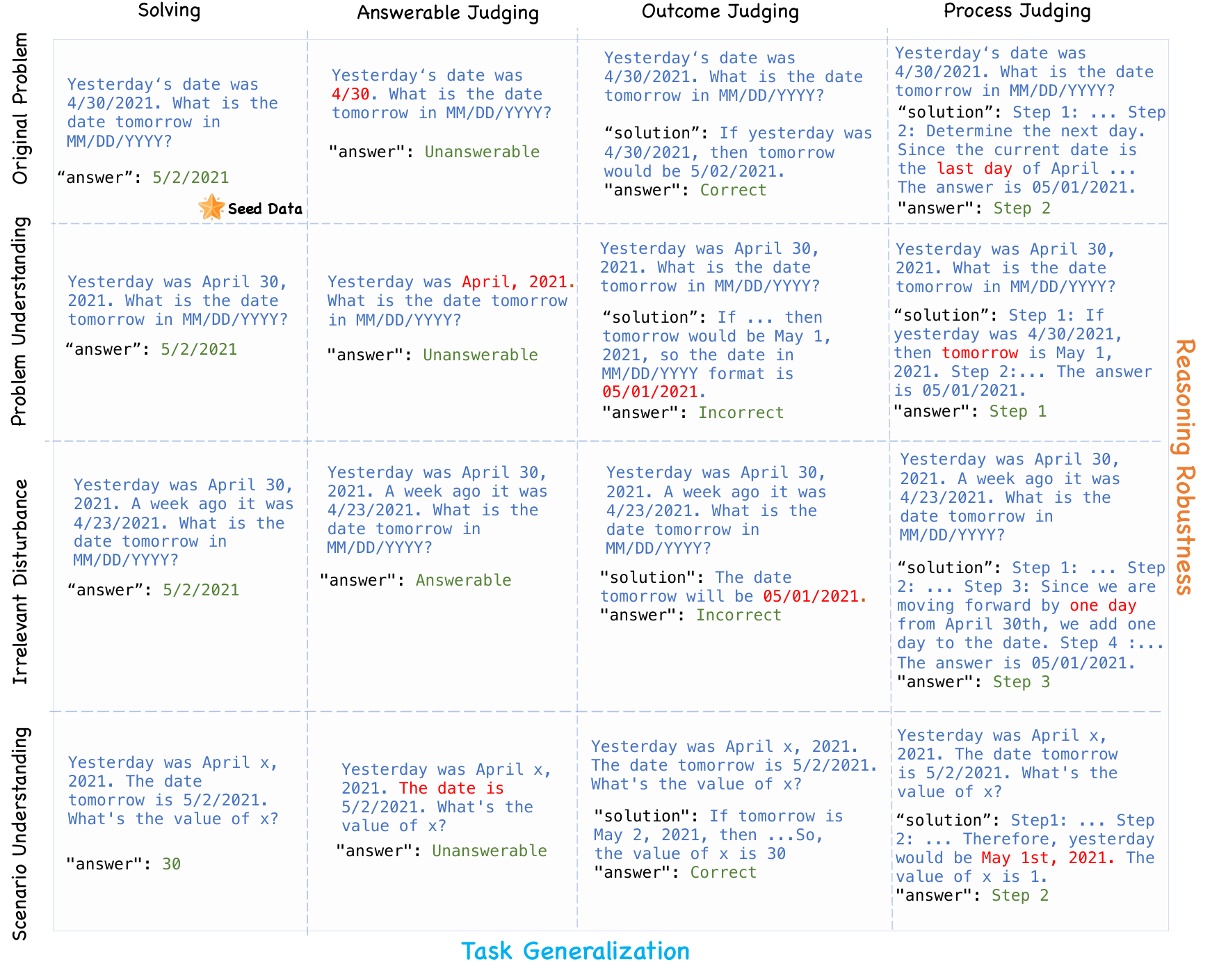}
    \caption{Case of \textsc{MathCheck} in  Date Understanding.}
    \label{fig:mathcheck_date}
\end{figure*}

\subsection{Date Understanding}
\label{sec:date}
To show that our proposed benchmark paradigm \textsc{MathCheck} can be adapted to other reasoning tasks beyond mathematical problems, we try to transform some representative reasoning task into \textsc{MathCheck} paradigm. We firstly apply it in commonsense reasoning, which requires LLMs to apply world knowledge to reason and solve problems. Specifically, we choose the date understanding task in Big-bench~\citep{srivastava2023beyond} since it is a wildly used task to measure commonsense reasoning ability~\citep{wei2022chain}. 

Figure~\ref{fig:mathcheck_date} shows the case of applying \textsc{MathCheck} to date understanding. Similar to mathematical reasoning, date understanding is a numerical reasoning task, therefore it can easily utilize variants of each unit in \textsc{MathCheck}. For example, in Irrelevant Disturbance, we can add some irrelevant date conditions to cause disturbance. In scenario understanding, we can ask for other variables in order to examine whether models have a comprehensive understanding of this date knowledge. This case demonstrates the high adaptability of \textsc{MathCheck} to commonsense reasoning task especially numerical reasoning.

\clearpage
\begin{figure*}[ht!]
    \centering
    \includegraphics[scale=0.47]{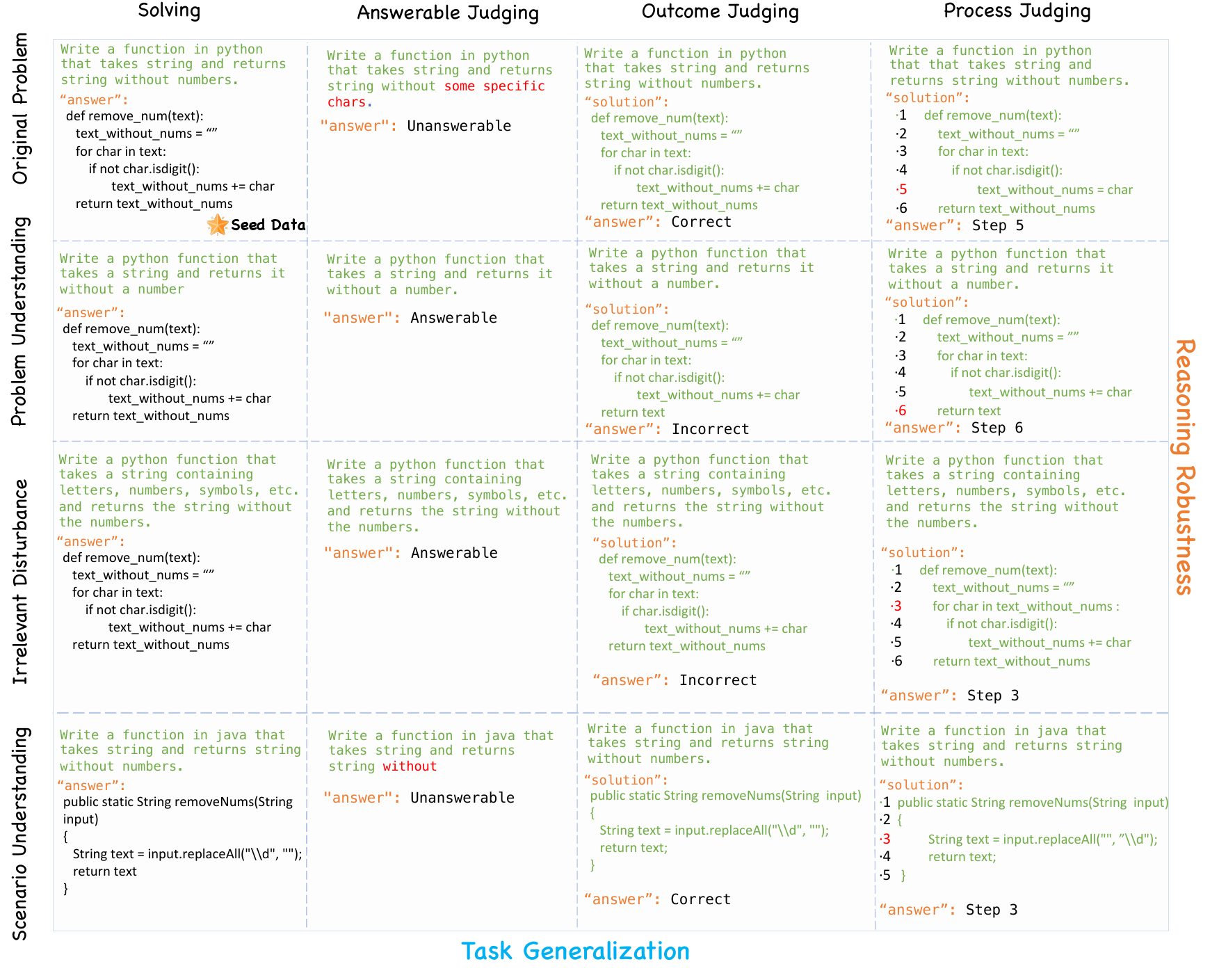}
    \caption{Case of \textsc{MathCheck} in Code Generation.}
    \label{fig:mathcheck_code}
\end{figure*}

\subsection{Code Generation}
\label{sec:code}
In addition to commonsense reasoning task, we would like to show the possibility of transforming \textsc{MathCheck} in some real-world reasoning tasks. Specifically, we choose the code generation task due to its high relevance to Text2Sql, agents and robotics. Figure~\ref{fig:mathcheck_code} demonstrates a case of applying \textsc{MathCheck} to code generation. Unlike numerical reasoning tasks, the adaptation of code generation needs to consider task relevance. For example, in Scenario Understanding, we can ask models to write the same function in other program languages (Python to Java in our case) in order to examine whether models have a comprehensive understanding of this function requirements. It shows that \textsc{MathCheck} have potential for real-world tasks such as agents and robotics application. Meanwhile, we encourage researchers to design more specific variants towards their reasoning task on \textsc{MathCheck} framework to test reasoning robustness and task generalization.

\clearpage

\section{Prompt List}
\label{Prompt List}

\lstset{
    backgroundcolor=\color[RGB]{245,245,245},
    breaklines=true,
    breakindent=0pt,
    basicstyle=\ttfamily\small,
    frame=trbl,
    frameround = tttt,
    captionpos=b  % 将 caption 放置在列表底部
}
\subsection{Evaluation Prompt}\label{secEvaPrompt}

\begin{lstlisting}[caption={Zero-shot Prompt of Problem Solving}]
You are an AI assistant that determines whether math problems are solved correctly. Answer the question. Finally give the answer in the format:
The answer is: ...

Question: [QUESTION]
Answer:
\end{lstlisting}

\begin{lstlisting}[caption={Zero-shot Prompt of Outcome Judging}]
You are an AI assistant that determines whether math problems are solved correctly. I will first give you a math problem and its solution, help me judge whether the final answer is correct or incorrect. Give your judgment between Correct or Incorrect. Finally summarize your answer in the format:
The answer is: ...

Question: [QUESTION]
Solution: [SOLUTION]
Judgement:
\end{lstlisting}

\begin{lstlisting}[caption={Zero-shot Prompt of Process Judging}]
You are an AI assistant that identify which step begins the error in solution. I will give you a math problem along with a wrong solution. Please help me identify the step where the errors begin. Finally give the wrong step in the format:
The answer is: Step i

Question: [QUESTION]
Solution: [SOLUTION]
Judgement:
\end{lstlisting}

\begin{lstlisting}[caption={Zero-shot Prompt of Answerable Judging}]
You are an AI assistant that determines whether math problems are answerable or unanswerable. Please analyze whether the question provides sufficient information to obtain an answer. Give your judgment between Answerable or Unanswerable. Finally summarize your answer in the format:
The answer is: ...

Question: [QUESTION]
Judgement:
\end{lstlisting}

\begin{lstlisting}[caption={Few-shot Prompt of Problem Solving}]
You are an AI assistant to help me solve math problems. Answer the question. Finally give the answer in the format: The answer is: ...
Follow the given examples and answer the question.

Question: Leah had 32 chocolates and her sister had 42. If they ate 35, how many pieces do they have left in total?
Answer: Step 1: Originally, Leah had 32 chocolates.
Step 2: Her sister had 42. So in total they had 32 + 42 = 74.
Step 3: After eating 35, they had 74 - 35 = 39.
The answer is 39.

Question: Jason had 20 lollipops. He gave Denny some lollipops. Now Jason has 12 lollipops. How many lollipops did Jason give to Denny?
Answer: Step 1: Jason started with 20 lollipops.
Step 2: Then he had 12 after giving some to Denny.
Step 3: So he gave Denny 20 - 12 = 8.
The answer is 8.


Question: [QUESTION]
Answer: 
\end{lstlisting}

\begin{lstlisting}[caption={Few-shot Prompt of Outcome Judging}]
You are an AI assistant that determines whether math problems are solved correctly. I will first give you a math problem and its solution, help me judge whether the final answer is correct or incorrect.

Give your judgment between Correct or Incorrect. Finally summarize your answer in the format: The answer is: ...
Follow the given examples and give your judgment.

Question: Leah had 32 chocolates and her sister had 42. If they ate 35, how many pieces do they have left in total?
Solution: Step 1: Originally, Leah had 32 chocolates.
Step 2: Her sister had 42. So in total they had 32 + 42 = 74.
Step 3: After eating 35, they had 74 - 35 = 39.
The answer is 39.
Judgment: Step 1 and Step 2 accurately calculate the total number of chocolates they both had originally.
Step 3 correctly calculates how many they have left after eating 35 chocolates.
The answer is: Correct.

Question: Jason had 20 lollipops. He gave Denny some lollipops. Now Jason has 12 lollipops. How many lollipops did Jason give to Denny?
Solution: Step 1: Jason started with 20 lollipops.
Step2: Then he had 12 after giving some to Denny.
Step3: So he gave Denny 20 + 12 = 8.
The answer is 32.
Judgment: Jason ended up with 12 lollipops after giving some to Denny, having started with 20. Therefore, the calculation to find out how many lollipops Jason gave to Denny should be:20 - 12 = 8. 
The answer is: Incorrect.


Question: [QUESTION]
Solution: [SOLUTION]
Judgement:
\end{lstlisting}

\begin{lstlisting}[caption={Few-shot Prompt of Process Judging}]
You are an AI assistant that identify which step begins the error in solution. I will give you a math problem along with a wrong solution. Please help me identify the step where the errors begin.

Finally give the wrong step in the format: The answer is: Step I
Follow the given examples and give your judgment.

Question: Leah had 32 chocolates and her sister had 42. If they ate 35, how many pieces do they have left in total?
Solution: Step 1: Originally, Leah had 32 chocolates.
Step 2: Her sister had 42. So in total they had 32 + 42 = 84.
Step 3: After eating 35, they had 84 - 35 = 49.\nThe answer is 49.
Judgment: The judgment of the given steps is as follows:
Step 1: Correctly states Leah's initial amount of chocolates.
Step 2: Incorrectly calculates the total number of chocolates both Leah and her sister had originally.
The answer is: Step 2.

Question: Jason had 20 lollipops. He gave Denny some lollipops. Now Jason has 12 lollipops. How many lollipops did Jason give to Denny?
Solution: Step 1: Jason started with 20 lollipops.
Step 2: Then he had 12 after giving some to Denny.
Step 3: So he gave Denny 20 + 12 = 8.
The answer is 32.
Judgment: The correct method to find out how many lollipops Jason gave to Denny would be to subtract the amount he had left from the amount he started with: 20 - 12 = 8. Thus, The reasoning error begins at Step 3. 
The answer is: Step 3.


Question: [QUESTION]
Solution: [SOLUTION]
Judgement:
\end{lstlisting}

\begin{lstlisting}[caption={Few-shot Prompt of Answerable Judging}]
You are an AI assistant that determines whether math problems are answerable or unanswerable. Please analyze whether the question provides sufficient information to obtain an answer.

Give your judgment between Answerable or Unanswerable. Finally summarize your answer in the format: The answer is: ...
Follow the given examples and give your judgment.

Question: Leah had 32 chocolates and her sister had 42. If they ate 35, how many pieces do they have left in total?
Judgment: The question provides all necessary information to perform the calculation. 
The answer is: Answerable.

Question: Jason had 20 lollipops. He gave Denny some lollipops. How many lollipops did Jason give to Denny?
Judgment: The question is not answerable as given. The reason is that there is insufficient information to determine the exact number of lollipops Jason gave to Denny.
The answer is: Unanswerable.

Question: [QUESTION]
Judgement:
\end{lstlisting}

\subsection{Data Generation Prompt}\label{secGenPrompt}

\begin{lstlisting}[caption={Prompt of Problem Understanding Rewriting}]
Your objective is to rewrite a given math question using the following perturbation strategy. The rewritten question should be reasonable, understandable, and able to be responded to by humans.

Perturbation strategy: Problem Understanding: It refers to transforming the original problem into a new problem that uses different wording or different sentence structures but does not change the solution of the original problem.

The given question: {QUESTION}
Answer of the given question: {ANSWER}

Please rewrite the question using the specified perturbation strategy while minimizing edits to avoid significant deviation in the question content.
It is important to ensure that the rewritten question has only one required numerical answer. You just need to print the rewritten question without answer.
The rewritten question:
Question: {QUESTION}
Answer: {ANSWER}
Given step: {STEP}
The rewritten answer:
\end{lstlisting}

\begin{lstlisting}[caption={Prompt of Irrelevant Disturbance Rewriting}]
Your objective is to rewrite a given math question using the following perturbation strategy. The rewritten question should be reasonable, understandable, and able to be responded to by humans.

Perturbation strategy: Irrelevant Disturbance: It involves introducing distracting conditions that have no impact on the final answer. These introduced conditions should be relevant to the topic of the original question and preferably include numerical values. However, the rewritten problem must maintain an identical solution to that of the original problem.

The given question: {QUESTION}
Answer of the given question: {ANSWER}

Please rewrite the question using the specified perturbation strategy while minimizing edits to avoid significant deviation in the question content.
It is important to ensure that the rewritten question has only one required numerical answer. You just need to print the rewritten question without answer.
The rewritten question:
Question: {QUESTION}
Answer: {ANSWER}
Given step: {STEP}
The rewritten answer:
\end{lstlisting}

\begin{lstlisting}[caption={Prompt of Unanswerable Question Rewriting}]
Your objective is to rewrite a given math question using the following perturbation strategy. The rewritten question should be reasonable, understandable, and able to be responded to by humans.

Perturbation strategy: Unanswerable question: It refers to eliminating a condition from the original question that is crucial for solving it while keeping the rest of the content unchanged. The rewritten problem should no longer have a valid answer, as it lacks the constraint that was removed.

The given question: {QUESTION}
Answer of the given question: {ANSWER}

Please rewrite the question using the specified perturbation strategy while minimizing edits to avoid significant deviation in the question content.
It is important to ensure that the rewritten question has only one required numerical answer. You just need to print the rewritten question without answer.
The rewritten question:
Question: {QUESTION}
Answer: {ANSWER}
Given step: {STEP}
The rewritten answer:
\end{lstlisting}

\begin{lstlisting}[caption={Prompt to Rewrite Question and Answer into a Declarative Statement}]
You are an AI assistant to help me rewrite question into a declarative statement when its answer is provided.
Follow the given examples and rewrite the question.

Question: How many cars are in the parking lot? The answer is 5.
Result: There are 5 cars in the parking lot.

Question: How many trees did the grove workers plant today? The answer is 6.
Result: The grove workers planted 6 trees today.

Question: If they ate 35, how many pieces do they have left in total? The answer is 39.
Result: They have 39 pieces left in total if they ate 35.

Question: How many lollipops did Jason give to Denny? The answer is 8.
Result: Jason gave 8 lollipops to Denny.

Question: How many toys does he have now? The answer is 9.
Result: He now has 9 toys.

Question: How many computers are now in the server room? The answer is 29.
Result: There are 29 computers now in the server room.

Question: How many golf balls did he have at the end of wednesday? The answer is 33.
Result: He had 33 golf balls at the end of Wednesday.

Question: How much money does she have left? The answer is 8.
Result: She has 8 money left.

Question: {QUESTION} The answer is {ANSWER}.
Result:
\end{lstlisting}

\begin{lstlisting}[caption={Prompt to Generate the Wrong Step}]
Following is a question and its correct solution. Rewrite the solution according to following requirements: (1) Do not change the format (2) Keep those steps before the given step unchanged (3) Make minor changes to the given step so that the reasoning of this step and subsequent steps are incorrect, resulting in an incorrect answer.

Question: {QUESTION}
Answer: {ANSWER}
Given step: {STEP}
The rewritten answer:
\end{lstlisting}

\clearpage

\section{Case Problems}

% Case GSM -  Group 54
\subsection{Case Problems in \textsc{MathCheck-GSM}. Problem Group ID: GSM-54}
\label{case_gsm}
\lstset{
    backgroundcolor=\color[RGB]{245,245,245},
    breaklines=true,
    breakindent=0pt,
    basicstyle=\ttfamily\small,
    frame=trbl,
    frameround=tttt,
    captionpos=b  % 将 caption 放置在列表底部
}
% -------------- Solving
\begin{lstlisting}[caption={Problem Solving - Original Problem}]
[Question]: Mike plays ping pong for 40 minutes. In the first 20 minutes, he scores 4 points. In the second 20 minutes, he scores 25% more points. How many total points did he score?
[Answer]: 9.0
\end{lstlisting}

\begin{lstlisting}[caption={Problem Solving - Problem Understanding}]
[Question]: During a 40-minute ping pong session, Mike scores 4 points in the initial half. In the latter half, he manages to increase his score by 25% compared to the first half. What is the total score Mike achieved in this session?
[Answer]: 9.0
\end{lstlisting}

\begin{lstlisting}[caption={Problem Solving - Irrelevant Disturbance}]
[Question]: Mike plays ping pong in a local tournament and decides to practice for 40 minutes before the first match. During his practice session, in the first 20 minutes, while intermittently checking his phone and hydrating, he manages to score 4 points. In the following 20 minutes, feeling more warmed up and despite a short break to adjust his paddle's grip tape, he scores 25% more points than in the first session. Considering these distractions, how many total points did Mike score in his 40-minute practice session?
[Answer]: 9.0
\end{lstlisting}

\begin{lstlisting}[caption={Problem Solving - Scenario Understanding}]
[Question]: Mike plays ping pong for 40 minutes.  In the first 20 minutes, he scores x points.  In the second 20 minutes, he scores 25% more points. He scored 9 total points. What is the value of unknown variable x?
[Answer]: 4.0
\end{lstlisting}

% -------------- Answerable Judging
\begin{lstlisting}[caption={Answerable Judging (\textit{Answerable}) - Original Problem}]
[Question]: Mike plays ping pong for 40 minutes. In the first 20 minutes, he scores 4 points. In the second 20 minutes, he scores 25% more points. How many total points did he score?
[Answer]: Answerable
\end{lstlisting}

\begin{lstlisting}[caption={Answerable Judging (\textit{Unanswerable}) - Original Problem}]
[Question]: Mike plays ping pong for minutes. In the first 20 minutes, he scores 4 points. In the second 20 minutes, his performance increases by 25%. How many total points did he score?
[Answer]: Unanswerable
\end{lstlisting}

\begin{lstlisting}[caption={Answerable Judging (\textit{Answerable}) - Problem Understanding}]
[Question]: During a 40-minute ping pong session, Mike scores 4 points in the initial half. In the latter half, he manages to increase his score by 25% compared to the first half. What is the total score Mike achieved in this session?
[Answer]: Answerable
\end{lstlisting}

\begin{lstlisting}[caption={Answerable Judging (\textit{Unanswerable}) - Problem Understanding}]
[Question]: During a 40-minute ping pong session, Mike scores points in the initial half. In the latter half, he manages to increase his score by 25% compared to the first half. What is the total score Mike achieved in this session?
[Answer]: Unanswerable
\end{lstlisting}

\begin{lstlisting}[caption={Answerable Judging (\textit{Answerable}) - Irrelevant Disturbance}]
[Question]: Mike plays ping pong in a local tournament and decides to practice for 40 minutes before the first match. During his practice session, in the first 20 minutes, while intermittently checking his phone and hydrating, he manages to score 4 points. In the following 20 minutes, feeling more warmed up and despite a short break to adjust his paddle's grip tape, he scores 25% more points than in the first session. Considering these distractions, how many total points did Mike score in his 40-minute practice session?
[Answer]: Answerable
\end{lstlisting}

\begin{lstlisting}[caption={Answerable Judging (\textit{Unanswerable}) - Irrelevant Disturbance}]
[Question]: Mike plays ping pong in a local tournament and decides to practice for 40 minutes before the first match. During his practice session, in the first 20 minutes, while intermittently checking his phone and hydrating, he manages to score points. In the following 20 minutes, feeling more warmed up and despite a short break to adjust his paddle's grip tape, he scores 25% more points than in the first session. Considering these distractions, how many total points did Mike score in his 40-minute practice session?
[Answer]: Unanswerable
\end{lstlisting}

\begin{lstlisting}[caption={Answerable Judging (\textit{Answerable}) - Scenario Understanding}]
[Question]: Mike plays ping pong for 40 minutes.  In the first 20 minutes, he scores x points.  In the second 20 minutes, he scores 25% more points. He scored 9 total points. What is the value of unknown variable x?
[Answer]: Answerable
\end{lstlisting}

\begin{lstlisting}[caption={Answerable Judging (\textit{Unanswerable}) - Scenario Understanding}]
[Question]: Mike plays ping pong for 40 minutes. In the first 20 minutes, he scores x points. In the second 20 minutes, he scores 25% more points. What is the value of unknown variable x?
[Answer]: Unanswerable
\end{lstlisting}

% -------------- Outcome Judging
\begin{lstlisting}[caption={Outcome Judging (\textit{Correct}) - Original Problem}]
[Question]: Mike plays ping pong for 40 minutes. In the first 20 minutes, he scores 4 points. In the second 20 minutes, he scores 25% more points. How many total points did he score?
[Solution]: 
Step 1: Find the number of points Mike scored in the second 20 minutes. To do this, take the number of points he scored in the first 20 minutes, which is 4 points, and increase this by 25%.
Step 2: To calculate 25% of 4, divide 4 by 4 which gives 1 (since 25% is the same as 1/4th).
Step 3: Add this 25% (which is 1 point) to the original 4 points to find out how many points he scored in the second 20 minutes. So, 4 points + 1 point = 5 points.
Step 4: Now, to find the total number of points Mike scored in the 40 minutes, add the points scored in the first 20 minutes to the points scored in the second 20 minutes.
So, 4 points (first 20 minutes) + 5 points (second 20 minutes) = 9 points.
The answer is 9 points.
[Answer]: Correct
\end{lstlisting}

\begin{lstlisting}[caption={Outcome Judging (\textit{Incorrect}) - Original Problem}]
[Question]: Mike plays ping pong for 40 minutes.  In the first 20 minutes, he scores 4 points. In the second 20 minutes, he scores 25% more points. How many total points did he score?
[Solution]: 
Step 1: Mike scores 4 points in the first 20 minutes.
Step 2: Mike scores 25% more points in the second 20 minutes.
Step 3: Mike's total points are the sum of his points in the first and second 20 minutes.
The answer is 28 points.
[Answer]: Incorrect
\end{lstlisting}

\begin{lstlisting}[caption={Outcome Judging (\textit{Correct}) - Problem Understanding}]
[Question]: During a 40-minute ping pong session, Mike scores 4 points in the initial half. In the latter half, he manages to increase his score by 25% compared to the first half. What is the total score Mike achieved in this session? 
[Solution]: 
Step 1: Calculate Mike's score in the first half, which is given as 4 points.
Step 2: Calculate the 25% increase in score for the second half. \\(25\\% \\text{ of } 4 = \\frac{25}{100} \\times 4 = 1\\).
Step 3: Add this increase to the initial score to find the score for the second half. \\(4 + 1 = 5\\).
Step 4: Combine the scores from both halves to get the total score. \\(4 \\text{ (first half) } + 5 \\text{ (second half) } = 9\\).
The answer is 9.
[Answer]: Correct
\end{lstlisting}

\begin{lstlisting}[caption={Outcome Judging (\textit{Incorrect}) - Problem Understanding}]
[Question]: During a 40-minute ping pong session, Mike scores 4 points in the initial half. In the latter half, he manages to increase his score by 25% compared to the first half. What is the total score Mike achieved in this session?
[Solution]:
Step 1: Mike scores 4 points in the initial half, so his score after 20 minutes is 20 minutes \\* 4 points/20 minutes = 8 points.
Step 2: In the latter half, Mike's score increases by 25% compared to the first half, so his score after 20 minutes is 8 points + (8 points \\* 25% increase) = 8 points + 2 points = 10 points.
Step 3: Therefore, Mike's total score after 40 minutes is 10 points + 4 points (from the initial half) = 14 points.
The answer is 14 points.
[Answer]: Incorrect
\end{lstlisting}

\begin{lstlisting}[caption={Outcome Judging (\textit{Correct}) - Irrelevant Disturbance}]
[Question]: Mike plays ping pong in a local tournament and decides to practice for 40 minutes before the first match. During his practice session, in the first 20 minutes, while intermittently checking his phone and hydrating, he manages to score 4 points. In the following 20 minutes, feeling more warmed up and despite a short break to adjust his paddle's grip tape, he scores 25% more points than in the first session. Considering these distractions, how many total points did Mike score in his 40-minute practice session? 
[Solution]: 
Step 1: Identify the points scored by Mike in the first 20 minutes, which is 4 points.
Step 2: Calculate the number of points scored in the next 20 minutes by finding 25% more than the points scored in the first session. 25% of 4 points is 1 (which is 0.25 * 4).
Step 3: Add the additional points to the first session's points to find the total points scored in the second session. So, 4 points + 1 point = 5 points in the second session.
Step 4: To find the total points scored in the 40-minute practice session, add the points from the first and second sessions together.
So, 4 points + 5 points = 9 points.\n\nThe answer is 9 points.
[Answer]: Correct
\end{lstlisting}

\begin{lstlisting}[caption={Outcome Judging (\textit{Incorrect}) - Irrelevant Disturbance}]
[Question]: Mike plays ping pong in a local tournament and decides to practice for 40 minutes before the first match. During his practice session, in the first 20 minutes, while intermittently checking his phone and hydrating, he manages to score 4 points. In the following 20 minutes, feeling more warmed up and despite a short break to adjust his paddle's grip tape, he scores 25% more points than in the first session. Considering these distractions, how many total points did Mike score in his 40-minute practice session?
[Solution]:
Step 1: Calculate the points scored in the first 20 minutes.\n4 points in 20 minutes = 2 points per minute
Step 2: Calculate the points scored in the second 20 minutes.\n25% more points than in the first session = 1.25 x 2 points/minute = 2.5 points/minute
Step 3: Calculate the total points scored in the 40-minute practice session.\n2 points/minute x 20 minutes = 40 points\n2.5 points/minute x 20 minutes = 50 points
Step 4: Add the points scored in the first and second sessions.\n40 points + 50 points = 90 points
The answer is 90 points.
[Answer]: Incorrect
\end{lstlisting}

\begin{lstlisting}[caption={Outcome Judging (\textit{Correct}) - Scenario Understanding}]
[Question]: Mike plays ping pong for 40 minutes. In the first 20 minutes, he scores x points. In the second 20 minutes, he scores 25% more points. He scored 9 total points. What is the value of unknown variable x?
[Solution]:
Step 1: Define the given information: Mike plays ping pong for 40 minutes, scores x points in the first 20 minutes, and scores 25% more points in the second 20 minutes.
Step 2: Since he scores 25% more points in the second 20 minutes, we can express the points scored in the second 20 minutes as \\(x + 0.25x\\).
Step 3: This can be simplified as \\(1.25x\\), representing the points scored in the second 20 minutes.
Step 4: Since he scored 9 points in total, we can write the equation:\n\\[x + 1.25x = 9\\]
Step 5: By simplifying the equation, we combine like terms:\n\\[2.25x = 9\\]
Step 6: To find the value of x, we divide both sides of the equation by 2.25:\n\\[x = \\frac{9}{2.25}\\]
Step 7: Simplify the division:\n\\[x = 4\\]
The answer is x = 4.
[Answer]: Correct
\end{lstlisting}

\begin{lstlisting}[caption={Outcome Judging (\textit{Incorrect}) - Scenario Understanding}]
[Question]: Mike plays ping pong for 40 minutes. In the first 20 minutes, he scores x points. In the second 20 minutes, he scores 25% more points. He scored 9 total points. What is the value of unknown variable x?
[Solution]:
Step 1: Let's first find out how many points Mike scored in the first 20 minutes.
Step 2: Since Mike scored x points in the first 20 minutes, we know that x/2 = 9 (since he scored 9 points in total).
Step 3: Now, we know that x = 2(9) = 18.
The answer is x = 18.
[Answer]: Incorrect
\end{lstlisting}

% -------------- Process Judging
\begin{lstlisting}[caption={Process Judging - Original Problem}]
[Question]: Mike plays ping pong for 40 minutes. In the first 20 minutes, he scores 4 points. In the second 20 minutes, he scores 25% more points. How many total points did he score?
[Solution]: 
Step 1: Find the number of points Mike scored in the second 20 minutes. To do this, take the number of points he scored in the first 20 minutes, which is 4 points, and increase this by 25%.
Step 2: To calculate 25% of 4, divide 4 by 4 which gives 1 (since 25% is the same as 1/4th).
Step 3: Add this 25% (which is 2 points) to the original 4 points to find out how many points he scored in the second 20 minutes. So, 4 points + 2 points = 6 points.
Step 4: Now, to find the total number of points Mike scored in the 40 minutes, add the points scored in the first 20 minutes to the points scored in the second 20 minutes. So, 4 points (first 20 minutes) + 6 points (second 20 minutes) = 10 points.
The answer is 10 points.
[Answer]: Step 3
\end{lstlisting}

\begin{lstlisting}[caption={Process Judging - Problem Understanding}]
[Question]: During a 40-minute ping pong session, Mike scores 4 points in the initial half. In the latter half, he manages to increase his score by 25% compared to the first half. What is the total score Mike achieved in this session?
[Solution]: 
Step 1: Calculate Mike's score in the first half, which is given as 4 points.
Step 2: Calculate the 25% increase in score for the second half. \\(4 \\div 25 = 0.16\\).
Step 3: Add this increase to the initial score to find the score for the second half. \\(4 + 0.16 = 4.16\\).
Step 4: Combine the scores from both halves to get the total score. \\(4 \\text{ (first half) } + 4.16 \\text{ (second half) } = 8.16\\).
The answer is 8.16.
[Answer]: Step 2
\end{lstlisting}

\begin{lstlisting}[caption={Process Judging - Irrelevant Disturbance}]
[Question]: Mike plays ping pong in a local tournament and decides to practice for 40 minutes before the first match. During his practice session, in the first 20 minutes, while intermittently checking his phone and hydrating, he manages to score 4 points. In the following 20 minutes, feeling more warmed up and despite a short break to adjust his paddle's grip tape, he scores 25% more points than in the first session. Considering these distractions, how many total points did Mike score in his 40-minute practice session?
[Solution]:
Step 1: Identify the points scored by Mike in the first 20 minutes, which is 4 points.
Step 2: Calculate the number of points scored in the next 20 minutes by finding 25% more than the points scored in the first session. 25% of 4 points is 1 (which is 0.25 * 4).
Step 3: Add the additional points to the first session's points to find the total points scored in the second session. So, 4 points + 1 point = 6 points in the second session.
Step 4: To find the total points scored in the 40-minute practice session, add the points from the first and second sessions together. So, 4 points + 6 points = 10 points.
The answer is 10 points.
[Answer]: Step 3
\end{lstlisting}

\begin{lstlisting}[caption={Process Judging - Scenario Understanding}]
[Question]: Mike plays ping pong for 40 minutes. In the first 20 minutes, he scores x points. In the second 20 minutes, he scores 25% more points. He scored 9 total points. What is the value of unknown variable x?
[Solution]: 
Step 1: Define the given information: Mike plays ping pong for 40 minutes, scores x points in the first 20 minutes, and scores 25% more points in the second 20 minutes.
Step 2: Since he scores 25% more points in the second 20 minutes, we can express the points scored in the second 20 minutes as \\(x + 0.25x\\).
Step 3: This can be simplified as \\(1.25x\\), representing the points scored in the second 20 minutes.
Step 4: Since he scored 9 points in total, we can write the equation:\n\\[x + 1.25x = 9\\]
Step 5: By simplifying the equation, we combine like terms:\n\\[2.25x = 9\\]
Step 6: To find the value of x, we divide both sides of the equation by 2.25:\n\\[x = \\frac{9}{2.25}\\]
Step 7: Simplify the division:\n\\[x = 5\\]\n\nThe answer is x = 5.
[Answer]: Step 7
\end{lstlisting}

\clearpage

% Case GEO -  Group 54
% ° - \\degree, ⊙ - \\odot, ∠ -\\angle
\subsection{Case Problems in \textsc{MathCheck-GEO}. Problem Group ID: GEO-15}
\label{case_geo}
\lstset{
    backgroundcolor=\color[RGB]{245,245,245},
    breaklines=true,
    breakindent=0pt,
    basicstyle=\ttfamily\small,
    frame=trbl,
    frameround=tttt,
    captionpos=b  % 将 caption 放置在列表底部
}

\begin{figure*}[h!]
    \centering
    \includegraphics[scale=0.72]{case_geo_diagram.jpg}
    \caption{Geometry diagram for geometry problems in group 15.}
    \label{fig:case_geo_diagram}
\end{figure*}

% -------------- Solving
\begin{lstlisting}[caption={Problem Solving - Original Problem}]
[Question]: As shown in the figure, the diameter CD of \\odot O crosses the midpoint G of chord EF, \\angle DCF = 20.0, then \\angle EOD is equal to ()\\degree
[Answer]: 40.0
\end{lstlisting}

\begin{lstlisting}[caption={Problem Solving - Problem Understanding}]
[Question]: In the circle with center O, diameter CD intersects the midpoint G of the chord EF, and the measure of angle DCF is 20 degrees. Determine the measurement of angle EOD in degrees.
[Answer]: 40.0
\end{lstlisting}

\begin{lstlisting}[caption={Problem Solving - Irrelevant Disturbance}]
[Question]: In the figure of circle O, the diameter CD intersects the midpoint G of the chord EF. The length of the chord EF is 7.5 cm, which is irrelevant to our angle measurements. The angle \\angle DCF is given to be 20.0 degrees. We need to calculate the angle \\angle EOD. What is the measure of this angle in degrees?
[Answer]: 40.0
\end{lstlisting}

\begin{lstlisting}[caption={Problem Solving - Scenario Understanding}]
[Question]: As shown in the figure, the diameter CD of \\odot O crosses the midpoint G of chord EF, \\angle DCF = x , \\angle EOD is equal to 40\\degree. What is the value of unknown variable x?
[Answer]: 20.0
\end{lstlisting}

% -------------- Answerable Judging
\begin{lstlisting}[caption={Answerable Judging (\textit{Answerable}) - Original Problem}]
[Question]: As shown in the figure, the diameter CD of \\odot O crosses the midpoint G of chord EF, \\angle DCF = 20.0, then \\angle EOD is equal to ()\\degree
[Answer]: Answerable
\end{lstlisting}

\begin{lstlisting}[caption={Answerable Judging (\textit{Unanswerable}) - Original Problem}]
[Question]: As shown in the figure, the diameter CD of \\odot O crosses chord EF, \\angle DCF = 20.0, then \\angle EOD is equal to ()\\degree
[Answer]: Unanswerable
\end{lstlisting}

\begin{lstlisting}[caption={Answerable Judging (\textit{Answerable}) - Problem Understanding}]
[Question]: In the circle with center O, diameter CD intersects the midpoint G of the chord EF, and the measure of angle DCF is 20 degrees. Determine the measurement of angle EOD in degrees.
[Answer]: Answerable
\end{lstlisting}

\begin{lstlisting}[caption={Answerable Judging (\textit{Unanswerable}) - Problem Understanding}]
[Question]: In the circle with center O, diameter CD intersects the midpoint G of the chord EF. Determine the measurement of angle EOD in degrees.
[Answer]: Unanswerable
\end{lstlisting}

\begin{lstlisting}[caption={Answerable Judging (\textit{Answerable}) - Irrelevant Disturbance}]
[Question]: In the figure of circle O, the diameter CD intersects the midpoint G of the chord EF. The length of the chord EF is 7.5 cm, which is irrelevant to our angle measurements. The angle \\angle DCF is given to be 20.0 degrees. We need to calculate the angle \\angle EOD. What is the measure of this angle in degrees?
[Answer]: Answerable
\end{lstlisting}

\begin{lstlisting}[caption={Answerable Judging (\textit{Unanswerable}) - Irrelevant Disturbance}]
[Question]: In the figure of circle O, the diameter CD intersects the midpoint G of the chord EF. The length of the chord EF is 7.5 cm, which is irrelevant to our angle measurements. We need to calculate the angle \\angle EOD. What is the measure of this angle in degrees?
[Answer]: Unanswerable
\end{lstlisting}

\begin{lstlisting}[caption={Answerable Judging (\textit{Answerable}) - Scenario Understanding}]
[Question]: As shown in the figure, the diameter CD of \\odot O crosses the midpoint G of chord EF, \\angle DCF = x , \\angle EOD is equal to 40\\degree. What is the value of unknown variable x?
[Answer]: Answerable
\end{lstlisting}

\begin{lstlisting}[caption={Answerable Judging (\textit{Unanswerable}) - Scenario Understanding}]
[Question]: As shown in the figure, the diameter CD of \\odot O crosses the midpoint G of chord EF, \\angle DCF = x. What is the value of unknown variable x?
[Answer]: Unanswerable
\end{lstlisting}

% -------------- Outcome Judging
\begin{lstlisting}[caption={Outcome Judging (\textit{Correct}) - Original Problem}]
[Question]: As shown in the figure, the diameter CD of \\odot O crosses the midpoint G of chord EF, \\angle DCF = 20.0, then \\angle EOD is equal to ()\\degree
[Solution]: 
Step 1: Identify that the angle DCF is an angle formed by a chord (CF) and a tangent (CD) at the point C.
Step 2: Recognize that the angle between a chord and a tangent at the point of contact is equal to the angle in the alternate segment, which is angle ECF in this case.
Step 3: Thus, the measure of angle ECF is also 20 degrees.
Step 4: Observe that the diameter CD passes through the center O of the circle and cuts the chord EF at its midpoint G.
Step 5: Recall that a line drawn from the center of the circle to the midpoint of a chord is perpendicular to the chord.
Step 6: Therefore, triangle EOG is a right-angled triangle at G.
Step 7: Understand that angle EOD is double the angle ECF because angle EOD is the angle at the center whereas angle ECF is at the circumference based on the Circle Theorem.
Step 8: Calculate angle EOD as 2 times angle ECF, which is 2 * 20 degrees.
Step 9: Perform the multiplication: 2 * 20 degrees = 40 degrees.
The answer is: Angle EOD measures 40 degrees.
[Answer]: Correct
\end{lstlisting}

\begin{lstlisting}[caption={Outcome Judging (\textit{Incorrect}) - Original Problem}]
[Question]: As shown in the figure, the diameter CD of \\odot O crosses the midpoint G of chord EF, \\angle DCF = 20.0, then \\angle EOD is equal to ()\\degree
[Solution]: 
Step 1: Notice that chord EF is bisected by diameter CD at point G.
Step 2: Recognize that \\angle DCF is an angle formed by the diameter CD and chord EF, so it is inscribed in the semicircle of \\odot O.
Step 3: Recall that an angle inscribed in a semicircle is a right angle; hence \\angle DCF = 90\\degree (not the given 20\\degree, which seems to be a mistake).
Step 4: Realize that \\angle EOD is the angle at the center of the circle, and it stands on the arc EF.
Step 5: Understand that the angle at the center of a circle is twice the angle at the circumference when both stand on the same arc.
Step 6: Since \\angle DCF is a right angle (90\\degree) and \\angle EOD is twice that angle at the center, \\angle EOD is 2 * 90\\degree.
Step 7: Calculate 2 * 90\\degree to get 180\\degree.
The answer is \\angle EOD is equal to 180\\degree.
[Answer]: Incorrect
\end{lstlisting}

\begin{lstlisting}[caption={Outcome Judging (\textit{Correct}) - Problem Understanding}]
[Question]: In the circle with center O, diameter CD intersects the midpoint G of the chord EF, and the measure of angle DCF is 20 degrees. Determine the measurement of angle EOD in degrees.
[Solution]: 
Step 1: Identify that the angle DCF is an angle formed by a chord (CF) and a tangent (CD) at the point C.
Step 2: Recognize that the angle between a chord and a tangent at the point of contact is equal to the angle in the alternate segment, which is angle ECF in this case.
Step 3: Thus, the measure of angle ECF is also 20 degrees.
Step 4: Observe that the diameter CD passes through the center O of the circle and cuts the chord EF at its midpoint G.
Step 5: Recall that a line drawn from the center of the circle to the midpoint of a chord is perpendicular to the chord.
Step 6: Therefore, triangle EOG is a right-angled triangle at G.
Step 7: Understand that angle EOD is double the angle ECF because angle EOD is the angle at the center whereas angle ECF is at the circumference based on the Circle Theorem.
Step 8: Calculate angle EOD as 2 times angle ECF, which is 2 * 20 degrees.
Step 9: Perform the multiplication: 2 * 20 degrees = 40 degrees.
The answer is: Angle EOD measures 40 degrees.
[Answer]: Correct
\end{lstlisting}

\begin{lstlisting}[caption={Outcome Judging (\textit{Incorrect}) - Problem Understanding}]
[Question]: In the circle with center O, diameter CD intersects the midpoint G of the chord EF, and the measure of angle DCF is 20 degrees. Determine the measurement of angle EOD in degrees.
[Solution]:
Step 1: Identify that chord EF is intersected at its midpoint G by the diameter CD; this implies that triangle CGF is isosceles with CG = GF.
Step 2: Recognize that angle DCF is given as 20 degrees and that angle DGF is also 20 degrees because CGF is an isosceles triangle and angle CGF equals angle DGF.
Step 3: Realize that angle EOF is twice angle DGF since angle EOF is the angle subtended at the center by chord EF, and angle DGF is the angle subtended at the circumference by the same chord (the Angle at the Center Theorem).
Step 4: Calculate angle EOF as 2 * angle DGF which is 2 * 20 degrees = 40 degrees.
Step 5: Note that angle EOD is twice angle EOF because EO is a diameter of the circle, and therefore angle EOD subtends a semicircle, making it a straight angle or 180 degrees.
Step 6: Divide 180 degrees by the angle EOF to find angle EOD. Since EOF is 40 degrees, angle EOD is 180 degrees / 40 degrees = 4.5 times EOF.
Step 7: Determine that there is a mistake in the previous step: angle EOF is part of angle EOD and not the other way around. Therefore, angle EOD consists of two EOF angles plus the straight angle DOF, which is 180 degrees.
Step 8: Sum up the angles to find angle EOD: angle EOF (40 degrees) + angle DOF (180 degrees) + angle EOF (40 degrees) which is 40 + 180 + 40.
Step 9: Add the angles to get the measurement of angle EOD: 40 + 180 + 40 = 260 degrees.
The answer is 260 degrees.
[Answer]: Incorrect
\end{lstlisting}

\begin{lstlisting}[caption={Outcome Judging (\textit{Correct}) - Irrelevant Disturbance}]
[Question]: In the figure of circle O, the diameter CD intersects the midpoint G of the chord EF. The length of the chord EF is 7.5 cm, which is irrelevant to our angle measurements. The angle \\angle DCF is given to be 20.0 degrees. We need to calculate the angle \\angle EOD. What is the measure of this angle in degrees?
[Solution]: 
Step 1: Recognize that \\angle DCF is an angle formed by intersecting a chord (EF) and a diameter (CD) of the circle.
Step 2: Acknowledge that an angle formed by a chord and a tangent (or in this case a diameter) at the point of contact is equal to the angle in the alternate segment, which means \\angle DCF is equal to angle \\angle EOF.
Step 3: Consider that CD is a diameter, so triangle COD is an isosceles triangle with CO = OD (both are radii of the circle).
Step 4: Recall that the base angles in an isosceles triangle are equal, which means \\angle CDO = \\angle DCO.\nStep 5: Since \\angle CDF is given as 20.0 degrees and \\angle CDO = \\angle DCO, each of these angles is also 20.0 degrees.
Step 6: Understand that \\angle EOD is the sum of \\angle EOF and \\angle FOD (since \\angle EOF and \\angle FOD are adjacent angles).
Step 7: Since \\angle EOF is equal to \\angle DCF, which is 20.0 degrees, then \\angle EOF is also 20.0 degrees.
Step 8: Calculate the measure of \\angle EOD by adding the measures of \\angle EOF and \\angle FOD, which are both 20.0 degrees.
Step 9: Perform the calculation: \\angle EOD = \\angle EOF + \\angle FOD = 20.0 degrees + 20.0 degrees.
Step 10: Conclude that \\angle EOD measures 40.0 degrees.
The answer is 40.0 degrees.
[Answer]: Correct
\end{lstlisting}

\begin{lstlisting}[caption={Outcome Judging (\textit{Incorrect}) - Irrelevant Disturbance}]
[Question]: In the figure of circle O, the diameter CD intersects the midpoint G of the chord EF. The length of the chord EF is 7.5 cm, which is irrelevant to our angle measurements. The angle \\angle DCF is given to be 20.0 degrees. We need to calculate the angle \\angle EOD. What is the measure of this angle in degrees?
[Solution]:
Step 1: Calculate the points scored in the first 20 minutes.\n4 points in 20 minutes = 2 points per minute
Step 2: Calculate the points scored in the second 20 minutes.\n25% more points than in the first session = 1.25 x 2 points/minute = 2.5 points/minute
Step 3: Calculate the total points scored in the 40-minute practice session.\n2 points/minute x 20 minutes = 40 points\n2.5 points/minute x 20 minutes = 50 points
Step 4: Add the points scored in the first and second sessions.\n40 points + 50 points = 90 points
The answer is 90 points.
[Answer]: Incorrect
\end{lstlisting}

\begin{lstlisting}[caption={Outcome Judging (\textit{Correct}) - Scenario Understanding}]
[Question]: As shown in the figure, the diameter CD of \\odot O crosses the midpoint G of chord EF, \\angle DCF = x , \\angle EOD is equal to 40\\degree. What is the value of unknown variable x?
[Solution]:
Step 1: Recognize that diameter CD of circle O passes through the center of the circle (point O) and bisects chord EF at point G.
Step 2: Since CD is a diameter, it creates a semicircle from E to F passing through C.
Step 3: Recognize that \\angle EOD is the angle subtended by the diameter and thus is a right angle (90\\degree) because it is an angle in a semicircle.
Step 4: Understand that triangle EOD is a right triangle with \\angle EOD being 90\\degree and \\angle EOF being half of \\angle EOD because O is the center of the circle and EOF is half of the circle (an arc of 180\\degree).
Step 5: Since \\angle EOD is 40\\degree as given, \\angle EOF must be half of that, so \\angle EOF = 40\\degree / 2 = 20\\degree.
Step 6: Notice that \\angle DCF and \\angle EOF are vertical angles and hence are equal. Therefore, x = \\angle DCF = \\angle EOF.
Step 7: Substitute the value from Step 5 into the equation for x, therefore x = 20\\degree.
The answer is x = 20\\degree.
[Answer]: Correct
\end{lstlisting}

\begin{lstlisting}[caption={Outcome Judging (\textit{Incorrect}) - Scenario Understanding}]
[Question]: As shown in the figure, the diameter CD of \\odot O crosses the midpoint G of chord EF, \\angle DCF = x , \\angle EOD is equal to 40\\degree. What is the value of unknown variable x?
[Solution]:
Step 1: Recognize that angle \\angle EOD lies on the diameter of the circle, which means it's a central angle and the arc ED it intercepts is 2x\\angle EOD = 2x40\\degree = 80\\degree.
Step 2: Since CD is a diameter, any angle formed on the circumference by diameter CD is a right angle (90\\degree). Hence, angles \\angle CFE and \\angle CED are 90 degrees each.
Step 3: The arc EG is half the circumference of the circle intercepted by angle \\angle CFE, so the measure of arc EG is 180\\degree since \\angle CFE is a right angle.
Step 4: Arc GF is the same as arc EG because G is the midpoint of EF, hence arc GF is also 180\\degree.
Step 5: The entire arc EF, which is the sum of arcs EG and GF, is 180\\degree + 180\\degree = 360\\degree - 80\\degree = 280\\degree since the circle is 360\\degree and we must subtract arc ED (80\\degree) which is not part of arc EF.
Step 6: The angle \\angle DCF is an inscribed angle that intercepts arc EF. The measure of an inscribed angle is half the measure of the arc it intercepts.
Step 7: Calculate the measure of angle \\angle DCF which is half of the measure of arc EF: x = 1/2 * 280\\degree = 140\\degree.
The answer is 140\\degree.
[Answer]: Incorrect
\end{lstlisting}

% -------------- Process Judging
\begin{lstlisting}[caption={Process Judging - Original Problem}]
[Question]: As shown in the figure, the diameter CD of \\odot O crosses the midpoint G of chord EF, \\angleDCF = 20.0, then \\angleEOD is equal to ()\\degree
[Solution]: 
Step 1: OE and OF are the radius of the circle, so OE=OF.
Step 2: G is the midpoint of EF, so EG=FG.
Step 3: Because triangle OGE and triangle OGF are congruent triangles.
Step 4: so the angle EOD is equal to the angle FOG.
Step 5: because the angle FOG is the outer angle of the triangle COF, the angle FOG=angle DCF+angle CFO.
Step 6: because OC=OF (they are all radius of circles), angle DCF=angle CFO.
Step 7: therefore the angle FOG=2* angle DCF.
Step 8: so the angle EOD = 3 * angle DCF.
Step 9: Given the angle DCF=20.0 , the angle EOD=60.0.
The answer is: Angle EOD measures 60.0 degrees.
[Answer]: Step 8
\end{lstlisting}

\begin{lstlisting}[caption={Process Judging - Problem Understanding}]
[Question]: In the circle with center O, diameter CD intersects the midpoint G of the chord EF, and the measure of angle DCF is 20 degrees. Determine the measurement of angle EOD in degrees.
[Solution]: 
Step 1: OE and OF are the radius of the circle, so OE=OF.
Step 2: G is the midpoint of EF, so EG=FG.
Step 3: Because triangle OGE and triangle OGF are congruent triangles.
Step 4: so the angle EOD is equal to the angle FOG.
Step 5: because the angle FOG is the outer angle of the triangle COF, the angle FOG=angle DCF+angle CFO.
Step 6: because OC=OF (they are all radius of circles), angle DCF=2*angle CFO.
Step 7: therefore the angle FOG=1.5* angle DCF.
Step 8: so the angle EOD = 1.5 * angle DCF.
Step 9: Given the angle DCF=20.0 , the angle EOD=30.0.
The answer is: Angle EOD measures 30.0 degrees.
[Answer]: Step 6
\end{lstlisting}

\begin{lstlisting}[caption={Process Judging - Irrelevant Disturbance}]
[Question]: In the figure of circle O, the diameter CD intersects the midpoint G of the chord EF. The length of the chord EF is 7.5 cm, which is irrelevant to our angle measurements. The angle \\angleDCF is given to be 20.0 degrees. We need to calculate the angle \\angleEOD. What is the measure of this angle in degrees?
[Solution]:
Step 1: OE and OF are the radius of the circle, so OE=OF.
Step 2: G is the midpoint of EF, so EG=FG.
Step 3: Because triangle OGE and triangle OGF are congruent triangles.
Step 4: so the angle EOD is equal to the angle FOG.
Step 5: because the angle FOG is the outer angle of the triangle COF, the angle FOG=angle DCF+angle CFO.
Step 6: because OC=OF (they are all radius of circles), angle DCF=angle CFO.
Step 7: therefore the angle FOG=angle DCF.
Step 8: so the angle EOD = angle DCF.
Step 9: Given the angle DCF=20.0 , the angle EOD=20.0.
The answer is: Angle EOD measures 20.0 degrees.
[Answer]: Step 7
\end{lstlisting}

\begin{lstlisting}[caption={Process Judging - Scenario Understanding}]
[Question]: As shown in the figure, the diameter CD of \\odot O crosses the midpoint G of chord EF, \\angleDCF = x , \\angleEOD is equal to 40\\degree. What is the value of unknown variable x?
[Solution]: 
Step 1: OE and OF are the radius of the circle, so OE=OF.
Step 2: G is the midpoint of EF, so EG=FG.
Step 3: Because triangle OGE and triangle OGF are congruent triangles.
Step 4: so the angle EOD is equal to the angle FOG.
Step 5: because the angle FOG is the outer angle of the triangle COF, the angle FOG=angle DCF+angle CFO.
Step 6: because OC=OF (they are all radius of circles), angle DCF=angle CFO.
Step 7: therefore the angle FOG=4* angle DCF.
Step 8: so the angle EOD = 4 * angle DCF.
Step 9: Given the angle EOD=40.0 , the angle DCF = x = 10.0.
The answer is x = 10 degrees.
[Answer]: Step 7
\end{lstlisting}

\end{document}